\documentclass{elsarticle}
\pdfoutput=1
\usepackage{graphicx} % Used to insert images
\graphicspath{{fig/}}
\usepackage{adjustbox} % Used to constrain images to a maximum size 
\usepackage{color} % Allow colors to be defined
\usepackage{enumerate} % Needed for markdown enumerations to work
\usepackage{geometry} % Used to adjust the document margins
\usepackage{amsmath} % Equations
\usepackage{amssymb} % Equations
\usepackage{eurosym} % defines \euro
\usepackage[mathletters]{ucs} % Extended unicode (utf-8) support
\usepackage[utf8x]{inputenc} % Allow utf-8 characters in the tex document
\usepackage[T1]{fontenc}

\usepackage{fancyvrb} % verbatim replacement that allows latex
\usepackage{grffile}
\usepackage{hyperref}
\usepackage{longtable} % longtable support required by pandoc >1.10
\usepackage{booktabs}  % table support for pandoc > 1.12.2
\usepackage{multirow}

\usepackage[figure,vlined,linesnumbered]{algorithm2e}

\usepackage{subcaption}

\definecolor{darkorange}{rgb}{.71,0.21,0.01}
\definecolor{darkgreen}{rgb}{.12,.54,.11}
\definecolor{darkred}{rgb}{.54,.1,.1}
\definecolor{gray}{gray}{0.45}
\definecolor{blue}{rgb}{0,.145,.698}

\newcommand{\R}[1]{{\color{black} #1}}

\hypersetup{
  breaklinks=true,  % so long urls are correctly broken across lines
  colorlinks=true,
  urlcolor=blue,
  linkcolor=darkorange,
  citecolor=darkgreen,
}
\DeclareMathOperator*{\argmax}{arg\,max}

\newcommand{\doi}[1]{\textsc{doi}: \href{http://dx.doi.org/#1}{\nolinkurl{#1}}}

\makeatletter
\def\ps@pprintTitle{%
 \let\@oddhead\@empty
 \let\@evenhead\@empty
 \def\@oddfoot{Accepted Manuscript. DOI: 10.1016/j.future.2020.01.011\hfill}%
 \let\@evenfoot\@oddfoot}
\makeatother

\begin{document}

\title{Implementing a GPU-based parallel MAX-MIN Ant System}
\author[uos]{Rafa{\l} Skinderowicz}
\ead{rafal.skinderowicz@us.edu.pl}

\address[uos]{University of Silesia in Katowice, Faculty of Science and
Technology,\\B\k{e}dzi\'nska 39, 41-205 Sosnowiec, Poland\\
\vspace{1em}
       {\rm
       \textcopyright 2020. This manuscript version is made available under the CC-BY-NC-ND 4.0 license http://creativecommons.org/licenses/by-nc-nd/4.0/
       }
}

\begin{abstract}
The MAX-MIN Ant System (MMAS) is one of the best-known Ant Colony Optimization
(ACO) algorithms proven to be efficient at finding satisfactory solutions
to many difficult combinatorial optimization problems.  The slow-down in
Moore's law, and the availability of graphics processing units (GPUs) capable of
conducting general-purpose computations at high speed, has sparked considerable
research efforts into the development of GPU-based ACO implementations.  In this
paper, we discuss a range of novel ideas for improving the GPU-based parallel MMAS
implementation, allowing it to better utilize the computing power offered by
two subsequent Nvidia GPU architectures.
Specifically, based on the weighted reservoir sampling algorithm
we propose a novel parallel implementation of the node selection procedure,
which is at the heart of the MMAS and other ACO algorithms.
We also present a memory-efficient implementation of another key-component
-- the tabu list structure -- which is used in the ACO's solution construction stage.
The proposed implementations, combined with the existing
approaches, lead to a total of six MMAS variants, which are evaluated on a set of
Traveling Salesman Problem (TSP) instances ranging from 198 to 3,795
cities. The results show that our MMAS implementation is competitive with
state-of-the-art GPU-based and multi-core CPU-based parallel ACO
implementations: in fact, the times obtained for the Nvidia V100 Volta GPU 
were up to 7.18x and 21.79x smaller, respectively. The fastest
of the proposed MMAS variants is able to generate over 1 million candidate solutions per second
when solving a 1,002-city instance.
Moreover, we show that, combined with the 2-opt local search heuristic, the
proposed parallel MMAS finds high-quality solutions for the TSP instances
with up to 18,512 nodes.

\end{abstract}

\begin{keyword}
parallel MAX-MIN Ant System \sep
weighted reservoir sampling \sep
Ant Colony Optimization \sep
GPU \sep
CUDA
\end{keyword}

\maketitle

\section{Introduction}
\label{sec:Introduction}

Ant Colony Optimization (ACO) is a population-based metaheuristic
inspired by the social behavior of ants~\cite{Dorigo2004}.
It has been successfully applied in solving many NP-hard problems,
including the Traveling Salesman Problem (TSP),
the Quadratic Assignment Problem,
and the Sequential Ordering Problem~\cite{Skinderowicz2017,Stutzle2000,Talbi2001}.
Being metaheuristic, the ACO does not guarantee 
finding an optimum solution, however it is often able to offer
satisfactory \emph{approximate} solutions within an acceptable
time compared to exact methods~\cite{Dorigo2018}.
However, even metaheuristics can be prohibitively time consuming
if faced with a large enough problem instance.
For this reason, a lot of research attention has been devoted both to 
improving the effectiveness of the ACO search process,
and to speeding up its execution~\cite{Pedemonte2011}.
The idea of applying GPU-based computing to the ACO
is an example of the latter.
In recent years, the use of graphics processing units (GPUs)
to speed up scientific computations has become commonplace.
This has been, in part, dictated by the slow-down in Moore's law,
and the progress made in the GPU architecture development, which has resulted
in more computing capacity, flexibility, and ease-of-use~\cite{Thompson2018}.
In fact, currently a significant proportion of the world's fastest supercomputers
is equipped with GPUs to accelerate their computations~\cite{top500}. 
Still, the efficient use of the computing capacity offered by GPUs
remains a difficult task~\cite{Jia2018}.

In an ACO, a population of agents (ants) construct, in parallel,
a set of solutions to the optimization problem begin tackled.
Unfortunately, the inherent parallel nature of the ACO
does not translate easily into an efficient GPU-based parallel
implementation~\cite{Cecilia2012,Dawson2013b,Skinderowicz2016}.
The difficulties arise partly from the fact that not all of the ACO
computations are independent, e.g., pheromone trail updates; as well as from the
computing restrictions inflicted by GPU architectures.
Although considerable research attention has been devoted to using GPUs to
speed up the ACO-based algorithms, both the improved computational capabilities
of successive generations of GPUs as well as novel algorithmic ideas, offer new
opportunities for greater progress.

In this paper, we present a GPU-based parallel implementation of the MAX-MIN
Ant System (MMAS), which is one of the best-performing ACO
variants for solving various optimization problems including
the production-distribution scheduling problem~\cite{Jia2019},
the blocks relocation problem~\cite{Jovanovic2019},
the routing and scheduling of home health care caregivers
problem~\cite{Decerle2019}, and the traveling purchaser
problem~\cite{Skinderowicz2018}.
Building on existing research we show how each of
the essential MMAS components can be parallelized to allow the efficient use of
the significant computing power offered by the current generation of GPUs.

The main contributions presented in this paper can be summarized as follows:
\begin{itemize}
\item
  We present a novel parallel implementation of the next node (proportional) selection
  procedure used in the MMAS and other ACO algorithms.
  The implementation is based on the weighted reservoir sampling algorithm and
  fits well within the parallel computing model of contemporary GPU architectures.
\item
  We present a novel, memory-efficient implementation of the tabu list structure
  used by the ACO solution construction process. The implementation
  allows for the better utilization of fast, but very size limited, shared memory 
  of GPUs.
\item
  Combining these novel ideas and the solutions from the literature, we 
  present a total of six MMAS variants and evaluate 
  their computational efficiency based on two subsequent generations
  of Nvidia GPUs, namely Pascal and Volta.
\item
  The computational evaluation based on a set of TSP instances ranging from 198
        to 3,795 cities
  shows that the proposed GPU-based MMAS is competitive with
  state-of-the-art GPU-based~\cite{Dawson2013,Dawson2013b,Cecilia2018}
  and multi-core CPU-based~\cite{Zhou2018} parallel ACO implementations.
  The times obtained were up to 7.18x and 21.79x smaller, respectively.
\item
  Acknowledging, that the ACO algorithms are typically paired with
  an efficient, problem-specific local search (LS) method, we 
  combine the proposed GPU-based MMAS with a parallelized 2-opt heuristic.
  The computational experiments that consider the TSP instances of up to 18,512
  nodes show that the proposed implementation is able to generate
  high-quality solutions, i.e., within 1\% from an optimum, in a relatively
  short time.
\end{itemize}

The remainder of this paper is organized as follows. In Section~\ref{sec:background}
we provide a brief description of the MAX-MIN Ant System and short piece on the
characteristics of
general-purpose GPU computations. Section~\ref{sec:related-work} summarizes the existing
work on applying GPUs in speeding up ACO computations, including the MMAS.
Our main ideas on the efficient implementation of the GPU-based MMAS are presented
in Section~\ref{sec:implementing-gpu-based-mmas}; while 
Section~\label{sec:experimental-results} presents an analysis of the computational 
experiments we conducted in order to evaluate the proposals, and compare them with the work
described in the literature.
Finally, we summarize our findings and provide a few ideas for future work
in Section~\ref{sec:conclusions}.

\section{Background}
\label{sec:Background}

\label{sec:background}

\subsection{MAX-MIN Ant System}
\label{sec:MAX_MIN_Ant_System}

The ACO metaheuristic belongs to a group of swarm-based metaheuristics (SBMs)
in which the problem-solving abilities are a result of the interactions of
simple information-processing units (agents)~\cite{Kennedy2006}.
The inspirations for the SBMs often come from biological systems including ant
colonies, swarms of bees, flocks of birds, and schools of fish, among
others~\cite{Engelbrecht2005}. Typically, the agents in the SBMs follow simple
rules and are given a certain degree of autonomy, e.g., in selecting the next
action to perform. The agents may also interact with each other, e.g., by
transferring data about the solutions found, or with the environment, e.g., by
depositing artificial pheromone trails that can be read by other agents
(indirect communication).
In addition to ACO, one of the most-successful SBMs are particle swarm
optimization (PSO) and artificial bee colony algorithms.

In the MMAS, a number of ants (agents) iteratively construct solutions to a
combinatorial optimization problem (COP)~\cite{Stutzle2000}. In this paper, we
focus on the TSP, following the existing research on parallel
ACO~\cite{Bai2009,Dawson2013,Dawson2013b,Delevacq2013,Zhou2017,Cecilia2018},
although, in principle, the ACO algorithms can be applied to any
COP~\cite{Dorigo2004}.

The TSP can be defined using a complete graph $G = (V, A)$, where $V$ is a set
of nodes numbered from 0 to $n-1$ ($n$ being the number of cities), and $A$ as
the set of edges (arcs) between the nodes, i.e., $E = \{ (i, j): i, j \in V, \, i
\ne j \}$. The set of nodes represents a set of cities to be visited by a
salesman, while each edge in $E$ corresponds to a road between a pair of
cities. Additionally, for every edge, $(i, j)$, a positive value, $d_{ij}$, is
given, which represents the distance (weight) between the cities, $i$ and $j$. If the TSP
is \emph{symmetric}, then $d_{ij} = d_{ji}$, otherwise the instance is
asymmetric (ATSP) and the distance from $i$ to $j$ might not equal the distance
from $j$ to $i$.
Typically, the distances between the nodes in graph $G$ satisfy the triangle
inequality but, in general, they may be arbitrary.
Solving the TSP is equivalent to finding the minimum weight Hamiltonian cycle
in graph $G$. In general, solving the Hamiltonian cycle is an NP-hard problem
and finding the optimum to the TSP is at least as difficult.
A comprehensive overview of both exact and approximate approaches to solving the TSP
can be found in the work of Applegate et al.~\cite{Applegate2007}.

In the ACO, the edges of graph $G$ correspond to the \emph{solution
components} from which the ants' solutions are being constructed.  Additionally,
for every edge, $(i, j) \in E$, there is an associated \emph{pheromone trail},
$\tau_{ij}(t)$, where $t$ denotes a (discrete) time.
In nature, the pheromones are chemical substances that some ant species
use as an indirect
medium of communication between individuals~\cite{Dorigo2004}.
In ACOs, including
the MMAS, the artificial pheromone trails are stored as positive real values,
and influence the probability of including the corresponding solution
components into the solutions being constructed by the ants. In the MMAS,
in contrast to other ACO algorithms, the values of the pheromone trails are
\emph{bounded} by limits: $\tau_{\rm min}$ and $\tau_{\rm
max}$~\cite{Stutzle2000}.

In the MMAS, an ant starts its solution construction process from an
initial node and in each of the subsequent steps it selects an edge (a solution
component) that connect its current node with one of the neighboring, as yet
unvisited, nodes.  The choice of an edge is probabilistic and depends on
so-called \emph{heuristic information} (available \emph{a priori}) and
the values of the pheromone trails.
Specifically, an ant, $k$, positioned at node $i$ selects an edge, $(i, j)$,
leading to node $j$ with the probability:\\
\begin{equation}
\label{eq:prob}
p_{ij}^k(t) = \frac{ [ \tau_{ij}(t) ]^\alpha [ \eta_{ij} ]^\beta }{ \sum_{l \in \mathcal{N}_i^k} [\tau_{il}(t)]^\alpha [ \eta_{il} ]^\beta }  \; \quad \textrm{if} \; j \in \mathcal{N}_i^k \, ,
\end{equation}\\
where $\tau_{ij}(t)$ is the value of the pheromone trail deposited on the edge,
$(i,j)$; $\eta_{ij}$ is the value of the heuristic information for the edge,
$(i, j)$; $\alpha$ and $\beta$ are parameters that control the relative
influence of the pheromone values and the heuristic information on the
probability; and, finally, $\mathcal{N}_i^k$, denotes the set of nodes
that neighbor $i$ to be visited the ant, $k$.
The heuristic information, $\eta_{ij}$, specifies how \emph{attractive} a
particular edge $(i,j)$ is, and in the case of the TSP, $\eta_{ij} = 1 /
d_{ij}$, makes an edge more attractive the shorter it is. 
This is based on the assumption that good quality solutions consist of edges
connecting nodes located near to each other~\cite{Helsgaun2009}.
Each ant stores the previously visited cities in a \emph{tabu list}, which
allows $\mathcal{N}_i^k$ to be computed, guaranteeing that only valid
Hamiltonian cycles are constructed.

%(fold) alg:mmas
\begin{algorithm}[h]
\SetKwInOut{Input}{Input}\SetKwInOut{Output}{Output}

\def\ant[#1]{\textrm{Ant}(#1)}
\def\route[#1]{\textit{route}_{\textrm{Ant}(#1)}}
\def\tabu[#1]{\textit{tabu}_{\textrm{Ant}(#1)}}
\def\iterbest{\textit{iter\_best}}
\def\globalbest{\textit{global\_best}}

Calculate pheromone trails limits: $\tau_{\rm{min}}$ and
$\tau_{\rm{max}}$ \\

Set pheromone trails values to $\tau_{\rm{max}}$ \label{alg:mmas:pherinit} \\

$\globalbest \leftarrow \emptyset$

\For{ $i \leftarrow 1$ \KwTo $\textit{\#iterations} $ }{
    \label{alg:mmas:main.loop.start}
    
    \For{ $j \leftarrow 0$ \KwTo $\textit{\#ants}-1$ }{
        $u \leftarrow \mathcal{U}\{0, n-1\}$ \quad \tcp{Select the first node randomly}
        $\route[j] \left[0\right] \leftarrow u$ \\
        Add $u$ to $\tabu[j]$
        
        \For(\tcp*[h]{Complete the solution (route)}){ $k \leftarrow 1$ \KwTo $n - 1$ }{
            $u \leftarrow \textrm{select\_next\_node}( \route[j] \left[k-1\right], \tabu[j] )$ \\
            $\route[j] \left[k\right] \leftarrow u$ \\
            Add $u$ to $\tabu[j]$
        }
    }
    
    $\iterbest \leftarrow \textrm{select\_shortest} \left(\route[0], \ldots, \route[\#ants-1] \right)$ 
    \label{alg:mmas:iter.best} \\
    
    \If{ $\globalbest = \emptyset$ {\bf or} $\iterbest$ {\rm is shorter than} $\globalbest$ }{
        $\globalbest \leftarrow \iterbest$ \\
        Update pheromone trails limits
        $\tau_{\rm{min}}$ and $\tau_{\rm{max}}$ using $\globalbest$
    }
    
    Evaporate pheromone according to $\rho$ parameter \\ 
    Deposit pheromone based on $\iterbest$  \label{alg:mmas:pher.deposition} \\
   \label{alg:mmas:main.loop.end}
}
\caption{The MAX-MIN Ant System.}
\label{alg:mmas}
\end{algorithm}
%(end)

The pseudocode for the MMAS is shown in Fig.~\ref{alg:mmas}.  At first, the initial
pheromone trail limits, $\tau_{\rm min}$ and $\tau_{\rm max}$, are computed
based on a solution constructed using the nearest neighbor heuristic.
Next, the pheromone trails values are set to $\tau_{\rm max}$
(line~\ref{alg:mmas:pherinit}).
In the main loop of the algorithm
(lines~\ref{alg:mmas:main.loop.start}--\ref{alg:mmas:main.loop.end}), each ant
constructs a complete solution to the problem starting from a randomly chosen
node.
After the solutions have been constructed, the \emph{iteration best}
solution is selected (line \ref{alg:mmas:iter.best}). If it is shorter than
the current \emph{global best} solution, it becomes the new global best, and
the trail limits are updated accordingly.
Finally, a pheromone update is performed.
This means lowering (evaporating) the values of the pheromone trails:
$\tau_{ij} \leftarrow {\rm max}\left( \rho \tau_{ij}, \tau_{\rm min} \right)$,
where $\rho$ is a parameter that controls the evaporation speed.
The values of the pheromone trails never drop below the minimum value,
$\tau_{\rm min}$, which ensures that all edges have a non-zero probability of
being selected even in the late stages of the algorithm's execution.
The pheromone trails' values only increase if they correspond to the components
of the current iteration's best solution (line~\ref{alg:mmas:pher.deposition}).
The values are increased  according to:
$\tau_{ij} \leftarrow {\rm min}\left( \tau_{ij} + \Delta_{ij}, \tau_{\rm max} \right)$, where
\[
\Delta_{ij} =
    \begin{cases}
    {\rm cost}( {\it iter\_best } )^{-1} \, , & {\rm if~~} (i, j) \in {\it iter\_best} \, , \\
    0 \, , & {\rm otherwise} \, .
    \end{cases}
\]
Increasing pheromone levels for the trails corresponding to the edges (solution
components) of good quality solutions, increases the probability that, in
subsequent iterations, the ants will choose these edges more often. This process
allows the algorithm to learn and construct higher quality solutions over
time~\cite{Dorigo2004}.

It is possible to use the current \emph{global best} value instead of the
iteration best~\cite{Stutzle2000}.  
It is worth noting, that in contrast to the Ant Colony System (ACS),
parallelization is made simpler because the MMAS \emph{lacks} a local pheromone
update~\cite{Skinderowicz2016}.
In fact, the pheromone trail values remain constant during the solution
construction phase, allowing a \emph{beforehand} computation of the product of
the pheromone trails and the heuristic values required by~Eq.~(\ref{eq:prob}).
This optimization is in common use as it reduces both the computation time,
and more importantly, the number of loads from the
memory~\cite{Cecilia2012,Dawson2013b}.
We also apply it in our work, storing the product in a matrix called
\emph{choice\_info}.

A single iteration of the MMAS has a $O(mn^2)$ time complexity, as each of the
$m$ ants constructs a complete solution to the problem in $n-1$ steps (assuming
that the starting node is chosen arbitrarily), $n$ being the size of the problem
instance.
Each step has a complexity of $O(n)$ as an ant has to move from the
current node to the next, chosen from up to $n-1$ unvisited nodes.
If candidate lists are used, the average complexity of the solution
construction process falls to $O(n \cdot \textit{cl}) = O(n)$, where
$\textit{cl}$ is a constant that denotes the size of the list.

\subsection{General-purpose computing using GPUs}
\label{sec:General_purpose_computing_on_GPUs}

Architectural differences between the CPUs and the GPUs allow the latter to
offer a higher computing power but often at the cost of reorganizing the
structure of the calculations to enable parallel execution~\cite{Kirk2016}.
For the sake of clarity of further discussion,
it is worth clarifying the distinction between \emph{parallelism} and
\emph{concurrency}. Assuming that the required computations were divided
into independent portions or tasks, parallel execution refers to the case in which
the available processing elements (cores) execute the tasks
\emph{at the same time}. Concurrency, on the other hand, is a more general
term also including the cases in which some of the computations
may not overlap in time. In the simplest case, concurrent computations
require only a single computing unit working in a time-shared manner.
\R{
In other words, concurrency allows to handle multiple computing tasks \emph{at
once}, while parallelism emphasises doing multiple computations \emph{at the same
time}.
}

GPUs allow for a high degree of parallelism as they typically contain several replicated
\emph{streaming multiprocessors} (SMs), each comprising a number of
\emph{processing elements} that share control units, a register file,
caches, and shared memory.
However, the number of computational tasks should typically exceed the number
of available processing elements. This is helpful in situations in which the
computations get stalled, e.g., while waiting for the data to be read from
memory. In such cases, it is possible to switch to
another task for which the necessary data are available.
Somewhat related is the description of GPUs as being \emph{throughput-oriented}
meaning that a large number of computations can be performed in a given
period of time, however, the speed of execution of individual computations could be low
compared with that of CPUs~\cite{Kirk2016}.

SMs schedule and execute hundreds of parallel threads in groups of 32 called
\emph{warps}.
The warps employ a model called a \emph{single-instruction, multiple-thread}
(SIMT), in which all threads start at the same program address and typically
execute the same instructions over different data (data-parallelism).
However, each thread has its own program counter and register state so its
execution may diverge from the other threads in the warp.
From the performance point of view, it is best to keep the
number of diverging executions as low as possible; although, in the newer
Nvidia architectures (Volta, Turing) the penalty paid is lower than previous
versions~\cite{Jia2018}.
It is also worth adding that the threads within a warp have access to
primitives allowing them to access each other's registers directly, i.e.,
without the need for accessing slower, shared memory.

Nvidia's Compute Unified Device Architecture (CUDA) provides a
\emph{programming model} that forms an abstract layer over the hardware
architecture~\cite{CUDA2018}.
The CUDA divides programs into CPU (host) and GPU (device) parts.
The host and the device have separate memory spaces, but a unified memory
extension exists in CUDA 6.0 (and newer versions) that allows the CPU and GPU
threads to store data in a shared address space.
A programmer can define functions, called \emph{kernels}, which are executed by
the GPU.
Each kernel is executed by a specified number of concurrent threads that are
divided into several \emph{blocks}, which in turn are organized into a
\emph{grid}.
Each thread-block is assigned to a single SM and can communicate through the
shared memory with the other threads within the same block.
The blocks are scheduled independently; hence, threads belonging to separate
blocks can only communicate using large but high-latency global memory.

Summarizing, each thread executing on the GPU has access to a memory hierarchy,
with the privately-accessed registers being the fastest, the shared memory
being a bit slower, a small but cached \emph{constant memory} also offering
relatively fast reads, and, finally, the global memory being the slowest.  L1
and L2 caches are also present but not directly accessible to the programmer.
The CUDA programming model assumes that a large number of threads (tens of
thousands) is executed concurrently to allow memory-related latencies to be
hidden.

\section{Related work}
\label{sec:Related_work}

\label{sec:related-work}

Being a population-based metaheuristic, the ACO naturally exhibits some degree
of parallelism~\cite{Dorigo2004}.
For example, there is no direct communication between the ants.
In fact, the ants cooperate \emph{indirectly} (stigmergy) by modifying the
values of the pheromone trails that correspond to the components of the problem
they select during the solution construction phase.
If the solutions are constructed quickly, as they are in the case of the TSP,
the frequent updates of the pheromone trails become problematic from the
parallelization point of view.
Overall, a lot of research has been devoted to the parallelization of the ACO,
especially for execution on multi-CPU systems aimed at both improving the
quality of the generated solutions, and shortening the execution
time~\cite{Randall2002, Chu2004, Manfrin2006}.
A good summary of the research was done by Pedemonte et
al.~\cite{Pedemonte2011}.
Although valuable, the CPU-based parallelization of the ACO is difficult to
transfer directly to the GPUs due to differences in the hardware architectures. 
In most of the approaches to CPU-based ACO parallelization, a coarse-grained
organization of computations is favored, with the multi-colony ACO being one of
the most efficient. 
The GPUs on the other hand, are \emph{throughput oriented} and contain
thousands of relatively simple processing elements.
Only recently has the
increasing number of CPU cores and  the availability of wide vector
instructions (e.g. AVX2) allowed for a more efficient, fine-grained
approach~\cite{Zhou2017}. For these reasons, the rest of the section will
focus on research that targets the GPU-based parallelization of the ACO.

The first attempts at using GPUs to speedup the ACO predate the CUDA
programming framework.
Catala et al.~\cite{Catala2007} presented a parallel ACO for solving the
Orienteering Problem, although some speed increases were reported, the
implementation was complicated as the authors had to use graphics generation
primitives to perform computations.
A similar programming approach was used by Wang et al.~\cite{Wang2009}, who
proposed a GPU-based MMAS.
The authors reported a modest speedup compared to a sequential, CPU-based MMAS
implementation.

\subsection{Task-based vs data-parallel approaches}

One of the first attempts to speed up the MMAS using the first generation of
general purpose GPUs was made by Bai et al.~\cite{Bai2009}.
This approach used multiple ant colonies with a single colony assigned to a single thread block, and each thread within the block assigned to an ant.
The distance matrix was stored in texture memory to facilitate cache memory.
Computational experiments showed the execution was around 2x
faster than a reference CPU implementation when solving the TSP.
This is an example of \emph{task-based} parallelism, as the threads are directly
mapped to the ants.
The problem with this approach is that it leads to \emph{warp-branching}, i.e.,
different threads within a thread-warp (using the Nvidia CUDA-based
terminology) are likely to take different execution paths as the ants follow
divergent paths, causing the remaining threads to wait.

A more efficient, \emph{data-parallel} approach was proposed by Cecilia et
al.~\cite{Cecilia2012}.
In that implementation, a single thread block is mapped to a single ant in the
Ant System (AS), that is, all threads within a thread block work on a single
solution to the problem.
This avoids the warp-branching present
in the task-based approach as the threads execute the same instructions but for
different data, i.e., nodes.
The authors considered block sizes of 16 to 1,024 threads.
The computational experiments done on the Nvidia Tesla C2050 GPU showed that
the best performance was obtained for 128 threads per block.
The work's most notable contribution was the introduction of the so-called
\emph{I-Roulette} (independent roulette) method for selecting, in parallel, the next
city to be visited by an ant.
This was the alternative to the proportional selection method, also known as
the Roulette Wheel Method (RWM), which was used originally.
In the I-Roulette method, the probabilities of selecting each of the unvisited
nodes (assigned to separate threads) are multiplied by random numbers, and the
node that has the maximum product is selected through a parallel reduction.
Although, the I-Roulette method did not produce the same results as the
sequential RWM, it was up to 2.36x faster.
The authors also considered two parallel pheromone update methods, in which the
simpler one used atomic instructions to allow the safe simultaneous
modification of memory by multiple threads.
Overall, the reported speedups were up to 20x faster compared to the sequential
implementation.

A valuable comparison between the task-based and data-parallel approaches can be
found in the work of Del{\'e}vacq et al.~\cite{Delevacq2013} who presented a
GPU-based parallel implementation of the MMAS for the TSP.
In the task based approach, each ant was assigned to a CUDA thread.
In the data-parallel approach, a whole thread-block was assigned to a single ant.
Moreover, the 3-opt local search for improving the ants' solutions was also
included in both approaches.
The data-parallel approach was significantly faster than the task-based one,
and up to 19.47x faster than the reference sequential implementation.
The inclusion of the 3-opt resulted in more modest speedups of up to 8.03x for
the data-parallel implementation.
The authors concluded that the 3-opt is not well suited to GPU architecture
as it has a low computation to memory access (reads and writes) ratio.

\subsection{Alternative Implementations of the RWM}

The I-Roulette method used by Cecilia et al.~\cite{Cecilia2012} was analyzed,
both experimentally and analytically, by Lloyd and Amos~\cite{Lloyd2017} who
concluded that it behaves in a \emph{qualitatively} different way to the RWM.
Specifically, it tends to increase the probability of selecting an edge with a
high pheromone value in cases where there are a large number of edges to choose
from and the majority of the pheromone is concentrated on one edge.
This results in a slight degradation in the quality of MMAS solutions for
TSP instances with more than 1,000 nodes.
On the other hand, there is also a slight improvement in the quality of the
solutions produced by the parallel ACS.

Another approach to speeding up the AS on GPUs was proposed by Uchida et
al.~\cite{Uchida2012}.
In this algorithm, the RWM was replaced by a method called the
\emph{stochastic trial}.
The stochastic trial utilizes a matrix that has its rows assigned to the nodes,
each containing the prefix sums of the selection probabilities for the
corresponding node.
During the solution construction phase, an ant located at the node, $i$, draws a
uniform number from the range $[0, 1]$ and checks if the cell from the $i$-th row
of the matrix corresponds to an unvisited node.
If it does, it is selected, otherwise the process is repeated a specified
number of times.
In the case of a failure, the next node is selected using a (slower) parallel RWM.
Together with a parallel pheromone
update method, the proposed algorithm was up to 43.47x faster than a sequential
AS executed on a CPU.

The data-parallel approach was also adopted by Dawson and
Stewart~\cite{Dawson2013b} who applied a GPU-based AS to the TSP.
The authors proposed a new, efficient parallel implementation of the RWM -- the
Double-Spin Roulette (DS-Roulette) method.
The DS-Roulette method consists of three stages.
In the first stage, all nodes are divided between four thread warps
(128 threads).
Within a warp, the yet to be visited nodes (cities) are determined,
and the threads perform a warp-level reduction of the selection probabilities
that correspond to the nodes. 
In the second stage, the reduced values are used by the RWM to select a winning
warp.
In the third stage, the winning warp draws a second random number and performs
a node selection from the assigned nodes.
The selected node becomes the final result of the DS-Roulette execution.
The DS-Roulette method avoids the block-level reduction, and its results are
closer to the results of the sequential RWM when compared to the proposals of
Uchida et al.~\cite{Uchida2012} and Cecilia et al.~\cite{Cecilia2012}.
Combined with a parallel pheromone update, the resulting algorithm was up to
82.3x faster than the CPU-based implementation when tested on the Nvidia GTX 580
GPU.

In subsequent work, Dawson and Stewart~\cite{Dawson2013} presented a parallel
AS in which candidate lists were used to speed up the node selection process.
By limiting the length of the candidate list to 32 they were able to exploit
the warp-level communication primitives provided by the CUDA to efficiently
implement the RWM.
Along with the tabu list compression
method by Uchida et al.~\cite{Uchida2012}, the resulting implementation was up
to 18x faster than its sequential counterpart.

\subsection{Recent Advancements}

In recent work, Cecilia et al.~\cite{Cecilia2018} discussed several aspects
of an efficient GPU-based AS implementation.
Specifically, they introduced a parallel implementation of the RWM that uses
scan and stencil patterns to efficiently select an unvisited node.
To further speed up the calculations, the authors applied a partial
synchronization between the warps within thread-blocks to create a
\emph{super-warp} comprised of two warps (64 threads).
Combined with the previous parallel pheromone update
methods~\cite{Cecilia2012}, the resulting implementation, being
\emph{state-of-the-art}, was up to 8x faster than the baseline version proposed
earlier.

The most recent proposals include work by Borisenko and
Gorlatch~\cite{Borisenko2018} who presented a GPU-based parallel implementation
of the ACO, combined with Simulated Annealing (SA) for the
optimization of the multi-product batch plants used, e.g., in the chemical
industry.
The proposed metaheuristic was able to quickly find near-optimal solutions,
making it a viable alternative to the exact, but very time-consuming,
branch-and-bound approach.
Another work, by Rey et al.~\cite{Rey2018}, discusses
an interesting hybrid-parallel ACO for solving the Vehicle Routing Problem
(VRP).
The first stage of the algorithm consists of the MMAS being executed on the GPU
and generating TSP routes which are then combined into the VRP solutions and
improved using LS procedures during the second stage being executed
on the CPU.
The GPU-based MMAS is also one of components in the recently proposed parallel
framework for the Multi-population Cultural Algorithm by Unold and
Tarnawski~\cite{Unold2017}.

One should also be aware that a lot of work has been done on efficient parallel
implementations of other SBMs including the PSO~\cite{Mussi2011} and bees
algorithm~\cite{Luo2014}.  A more general summary can be found in the work of
Tan and Ding~\cite{Tan2016}.

\section{Implementing a GPU-based MMAS}
\label{sec:Implementing_GPU_based_MMAS}

\label{sec:implementing-gpu-based-mmas}

In this section we discuss how the MMAS can be parallelized in order to achieve
efficient execution on GPUs.
We devote most of our attention to the solution construction phase of the
algorithm which is its most time-consuming part.

\subsection{Tabu implementation}
\label{sec:Tabu_implementation}

\label{sec:tabu}

In order to calculate the probabilities defined by Eq.~(\ref{eq:prob}) it is
necessary to determine the set of nodes yet to be visited by an ant.
In the Ant System, the MMAS, and other ACO algorithms, the nodes already
visited by the ant are typically stored in a data structure called a \emph{tabu
list}~\cite{Dorigo1996}.
If the nodes are added in the order in which they are
visited by the ant, then the tabu list comprises a partial solution to the
problem, which, at the end of the construction phase, becomes the complete
solution.
It is
worth noting, that in order to calculate the probabilities given by
Eq.~(\ref{eq:prob}) it is necessary to determine the, as yet, unvisited nodes,
however, the relative order in which they are considered is not important.
In fact, we want the tabu list to fit into the fast but small shared memory of
the GPUs' SMs.
Hence, we can generalize the notion of the tabu list (or \emph{tabu} for short
-- to avoid confusion) to any data structure that provides the following
operations:
\begin{itemize}
\item \texttt{mark}($v$) -- marks node $v$ as \emph{visited};
\item \texttt{is\_visited}($v$) -- returns \texttt{true} if the node $v$ has
already been visited by an ant, or otherwise \texttt{false};
\item \texttt{length}() -- returns a number that is equal to or \emph{greater}
than the number of yet unvisited nodes;
\item \texttt{get\_candidate}($i$) -- where $i \in \{0, 1, \ldots,
\texttt{length}() - 1\}$, returns either an unvisited node $u, \; u \in V$, or
a special \emph{sentinel} value $s, \; s \notin V$; we assume that if
\texttt{get\_candidate}() is executed for every $i \in \{0, 1, \ldots,
\texttt{length}() - 1\}$ then the returned set of values contains all the nodes
to be visited by an ant.
\end{itemize}
The non-obvious definition of \texttt{get\_candidate}() allows the tabu list to
be implemented using a \emph{bitmask}.
The general scheme for accessing the set of nodes to be visited using the
presented operations is shown in Fig.~\ref{alg:tabu}.

%(fold) alg:tabu
\begin{algorithm}[h]
 l $\leftarrow$ tabu.length()\;
 \For{$i \leftarrow 0$ \KwTo $l-1$}{
     $v \leftarrow $ tabu.get\_candidate(i)\;
     \If{ $v \ne $ sentinel }{
       $v$ can be processed\;
     }
 }
 \caption{
     General scheme for processing the tabu using the generalized scheme (see
     Sec.~\ref{sec:Tabu_implementation})
 }
\label{alg:tabu}
\end{algorithm}
%(end)

It is worth emphasising that the tabu is at the core of the MMAS and other ACO
algorithms, and its implementation is important for the efficiency of the whole
algorithm. 
In case of GPUs, the tabu should be stored in the \emph{shared}
(local) memory so that it can be accessed quickly~\cite{Cecilia2012}.
Unfortunately, the size of the shared memory available to each thread block is
very limited -- usually only 48kB to 96kB in the last few generations of Nvidia
GPUs~\cite{Jia2018}.
Hence, both the time and space complexity of the tabu are important.

\begin{figure}[h]
\centering
\includegraphics[]{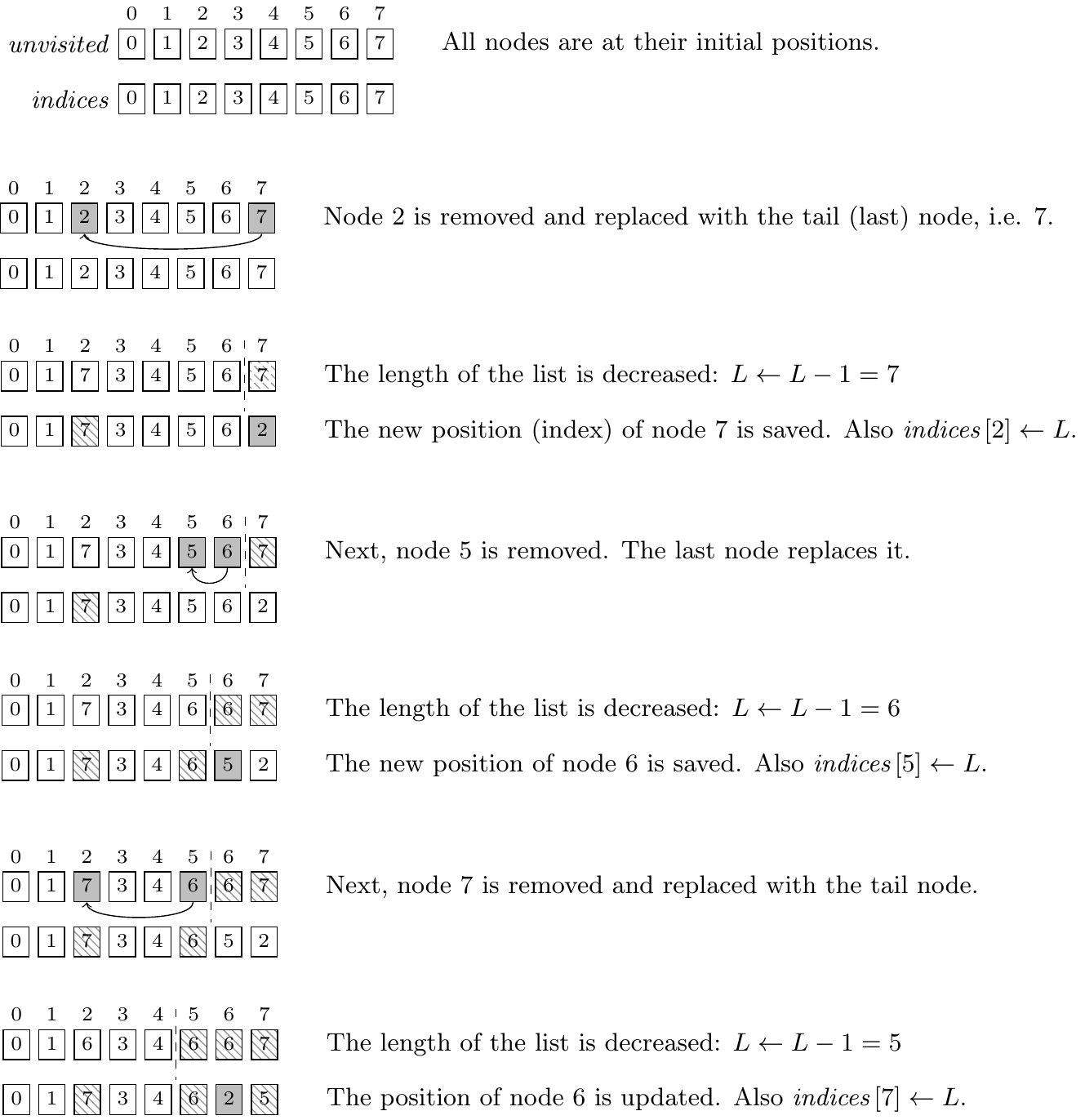}
\caption{An example showing the subsequent removal of nodes 2, 5, and 7 from the
LC tabu, which contains nodes 0 to 7. The dashed line marks the end of the nodes
list.}
\label{fig:compressed-tabu-example}
\end{figure}

A simple linked list is sufficient to implement all of the tabu operations,
however accessing an arbitrary element in the list has
$O(n)$ time complexity, with $n$ being the number of nodes.
A more efficient implementation, known as the \emph{tabu with list compression}
(LC), has been described by Dawson and Stewart~\cite{Dawson2013}, who applied
the \emph{list compression method} first proposed by~Uchida et
al.~\cite{Uchida2012}.
The data structure consists of two single
dimensional arrays of integers of size $n$, and a variable $L$.
For the sake of simplicity, lets denote them by $\textit{unvisited}$ and
$\textit{indices}$.
The first one, stores the list of $L$ nodes to be visited by an ant, while the
second stores the indices of every node in the first list.
For example, if $\textit{unvisited}[i] = v$ then $\textit{indices}[v] = i$ (we
assume that the nodes are denoted by numbers from 0 to $n-1$).
At the start of the construction process, both arrays contain a sequence of $n$
consecutive integers from $0$ to $n-1$, where $n$ is the number of nodes and
$L=n$.
In 
subsequent steps, if a node, $v = \textit{unvisited}[i]$, is being visited, then
the $\texttt{mark}(v)$ operation involves the updates: $\textit{unvisited}[i]
\leftarrow \textit{unvisited}[L-1]$, i.e., the visited node is replaced by the
last one, and $\textit{indices}\big[ \textit{unvisited}[i] \big] \leftarrow i$,
is followed by $L \leftarrow L-1$.
It can be seen that, if $\textit{indices}[u] \ge L$ then the node, $u$, has
already been visited by the ant.
Figure~\ref{fig:compressed-tabu-example} shows an example of how the LC works.

%(fold) fig:compressed-tabu-example
%(end)

The LC allows all the tabu operations to be performed in constant
time.
The major disadvantage is the necessity for storing two $n$-element arrays in
the memory.
On the other hand, the \textit{indices} array allows the
order in which the nodes were visited (in reverse order) to be recovered.
For example, if we consider the case shown in
Fig~\ref{fig:compressed-tabu-example}, nodes 2, 5 and 7 were already
visited, and the corresponding values in the \emph{indices} array are equal to 7, 6
and 5. By subtracting each value from $n-1$ we get $n-1-7=0$, $n-1-6=1$ and
$n-1-5=2$, respectively, which is exactly the order in which the nodes were
visited.

By analyzing the LC tabu, we notice that it is possible to implement the
tabu using only one list, \textit{entries}, of length $n$, and a variable,
$L$.
The trick is to divide the $\textit{entries}$ (logically) into two parts.
The first (left) part consists of the first $L \; (L \le n)$ entries and
contains the list of $L$ distinct nodes to be visited, i.e., $\textit{entries} \left[ i \right] = u \; ,
i < L$ if, and only if, node $u$ is as yet unvisited.
The second (right) part comprises entries at positions from $L$ up to $n-1$.
It is used to store indices for the nodes \emph{yet to be visited} that were
relocated to the left part, or the \emph{sentinel} value of $n$ for the nodes
that have already been \emph{visited}, i.e., are not in the left part.

\begin{figure}[h]
\centering
\includegraphics[]{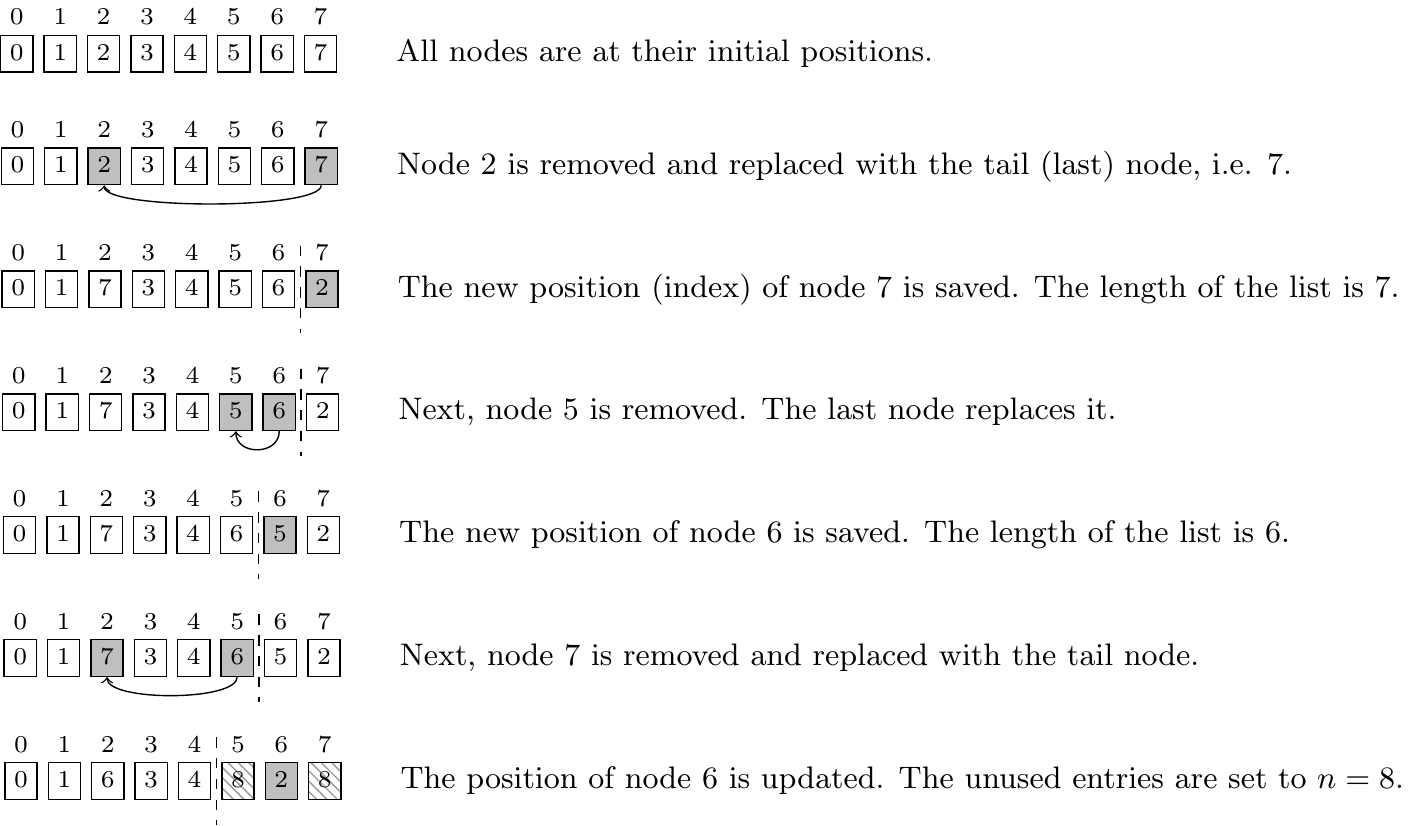}
\caption{
    An example showing the subsequent removal of nodes 2, 5, and 7 from a CT
    containing nodes 0 to 7.  The dashed line marks the end of the nodes
    list. The entries to the right of the dashed line are used to store
    positions (indices) for the nodes relocated to the left.
}
\label{fig:compact-tabu-example}
\end{figure}

Initially, the \textit{entries} array contains consecutive numbers from 0 to
$n-1$, denoting the unvisited nodes.
In subsequent steps, if a node, $u$, is visited, one of two cases is possible:
either $u < L$ or $u \ge L$.
In the first case, the node $u$ is at its initial position, i.e., $\textit{entries}
\left[ u \right] = u$.
In the second case, the node has been relocated to the left
part, and the value $i_u = \textit{entries} \left[ u \right]$ denotes the
\emph{index} at which the node is currently located, i.e.,  $\textit{entries}
\left[ i_u \right] = u$.
If $i_u = L-1$ then $u$ is at the end of the list,
and it is enough to set $\textit{entries} \left[ i_u \right] \leftarrow n$ to
mark that node $u$ has been visited.
Otherwise, $i_u < L-1$ and the last element, $t = \textit{entries}\left[ L-1
\right]$, of the list replaces it:
$\textit{entries} \left[ i_u \right] \leftarrow t$.
The new position of $t$ is
saved: $\textit{entries} \left[ t \right] \leftarrow i_u$.
It is worth noting that this scheme allows for checking in $O(1)$ time whether
node $u$ was visited,
simply by checking whether $\textit{entries} \left[ u \right] > u$.
We will refer to this tabu implementation as the \emph{compact tabu} (CT).
An example showing the removal of three nodes from a CT that contains eight
nodes is shown in Fig.~\ref{fig:compact-tabu-example}.

%(fold) fig:compact-tabu-example
%(end)

Even more memory efficient implementation can be achieved if a \emph{bitmask}
of length $n$ is used to mark the visited /unvisited state of each node.
In that case, the \texttt{mark} and \texttt{is\_visited} tabu operations can be
performed in $O(1)$ time.
However, accessing the unvisited nodes (Fig.~\ref{alg:tabu}) requires checking
the state of $n$ bits, while the two previous tabu implementations provide
\emph{direct} accesses to the \emph{unvisited} nodes. 
For the sake of completeness, we will refer to this
tabu as the \emph{bitmask tabu} (BT).
A summary of the presented tabu implementations is shown in
Tab.~\ref{tab:tabu-cmp}.

%(fold) tab:tabu-cmp
\begin{table}[]
\footnotesize
\centering
\caption{
    A summary of the considered tabu implementations. We assume that the nodes
    are denoted as numbers 0 to $n-1$, where $n < 2^{16}$, i.e. a node can
    fit into an unsigned 16-bit variable.
}
\label{tab:tabu-cmp}
\begin{tabular}{@{}lrrr@{}}
\toprule
\multirow{2}{*}{Tabu characteristic}  & \multicolumn{3}{c}{Tabu implementation}                                                \\ \cmidrule(l){2-4} 
                                      & \multicolumn{1}{c}{TLC} & \multicolumn{1}{c}{CT} & \multicolumn{1}{c}{BT}              \\ \midrule
Testing if node was visited (\texttt{is\_visited})          & $O(1)$                  & $O(1)$                 & $O(1)$                              \\ \cmidrule(r){1-1} \cmidrule(l){2-4}
Marking node as visited (\texttt{mark})              & $O(1)$                  & $O(1)$                 & $O(1)$                              \\ \cmidrule(r){1-1} \cmidrule(l){2-4}
\begin{tabular}[c]{@{}l@{}}Number of calls to \texttt{get\_candidate} \\ to get all $k \; (k \le n)$ unvisited nodes\end{tabular}        &        $k$         & $k$                    & $n$                           \\ \cmidrule(r){1-1} \cmidrule(l){2-4}
Required memory in bytes              & $4n$                    & $2n$                   & $\left\lceil n/8 \right\rceil$ \\ \bottomrule
\end{tabular}
\end{table}
%(end)

\subsection{Next node selection}
\label{sec:Next_node_selection}

\label{sec:next.node.selection}

The procedure for the selection of the next node by an ant during the solution
construction phase has the biggest impact on the performance of the MMAS and other
ACO algorithms~\cite{Cecilia2012,Uchida2012}.
Equation~(\ref{eq:prob}) defines the probabilities for selecting each of the
unvisited nodes.
The sequential version of the procedure has a simple and efficient
implementation, often referred to as the \emph{roulette wheel method} (RWM).
The RWM takes $O(n)$ time, where $n$ is the number of nodes to choose from.

\subsubsection{Parallel RWM}
\label{sec:Parallel_RWM}

The RWM is a typical example of an algorithm that has a simple and efficient
sequential implementation but is difficult to parallelize
effectively~\cite{Cecilia2012}.  The difficulties reside in the dependencies
between the subsequent computations of the RWM (e.g., the summation of the
pheromone and heuristic information products [weights in short]), searching for
a "winning" node based on a randomly drawn value.
It is not surprising that
multiple alternative RWM implementations have been proposed in the literature,
including the I-Roulette method by Cecilia et. al~\cite{Cecilia2012}, the DS-Roulette
by Dawson and Stewart~\cite{Dawson2013b}, and the stochastic trial by Uchida et.
al~\cite{Uchida2012}.
Although these methods allow for an efficient parallel
execution, they are \emph{qualitatively} different from the sequential
version~\cite{Lloyd2017}.
Only recently has Cecilia et. al~\cite{Cecilia2018}
proposed the parallel \emph{SS-Roulette} method, which is essentially a parallel
version of the RWM, i.e., it offers the same quality of results as the
sequential implementation.
This was possible mainly due to the increasing
computational capacity of GPU architectures, and also improvements
on the software side, e.g., the CUDA toolkit.

\begin{figure}[h]
\centering
\includegraphics[]{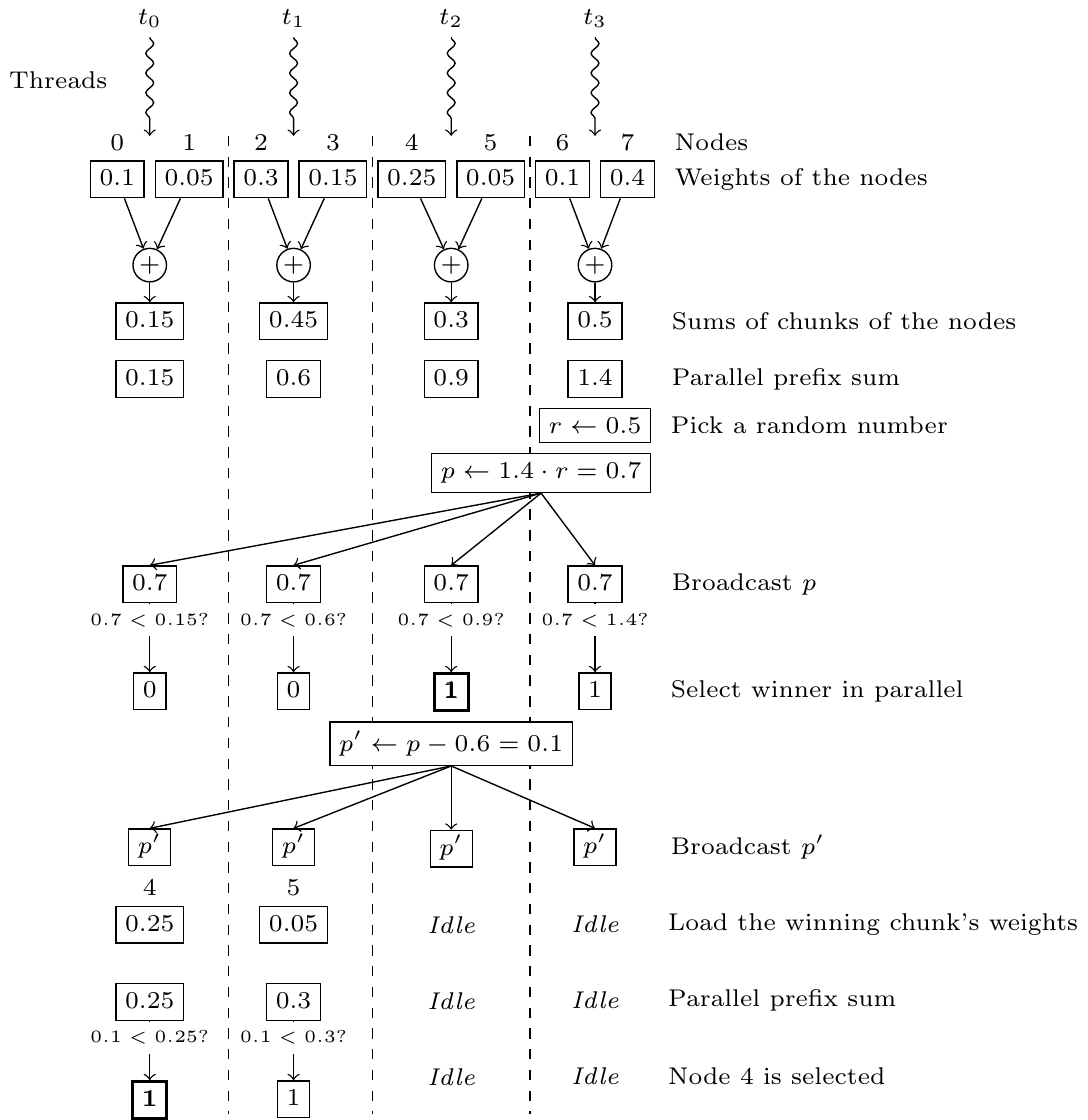}
\caption{An example showing the execution of the parallel RWM.}
\label{fig:par-rwm}
\end{figure}

Following the description of the SS-Roulette method (as the source code is not
available), we have implemented a parallel version of the RWM.
Figure~\ref{fig:par-rwm} shows how our implementation of the parallel RWM
(PRWM) works.
First, each thread that is executing the PRWM is assigned a chunk of the
unvisited nodes, for which it computes a sum of the corresponding weights.
Next, a prefix sum of the chunks' sums is computed in parallel.
Following this, the last thread draws a uniform random number and multiplies it
by the total.
The resulting value is broadcast to all threads so that the \emph{winning}
chunk of nodes can be selected.
If the chunk contains more than one node, then it is necessary to locate the
selected node within the chunk.
This process again involves the calculation of the prefix sums of the nodes'
weights but this time only for the nodes within the chunk.
This can also be done in parallel by splitting the chunk's nodes across all
threads.
In general, this process may require up to $\left \lceil \log_p n
\right \rceil$ stages, where $n$ is the number of nodes and $p$ is the number
of processors.
For example, if $n = 1024$ and $p=32$, then in the first stage
each of the 32 threads processes 32-node chunks, and in the second (final)
stage each thread is assigned one out of the 32 nodes from the chunk selected
during the first stage.
It is worth noting that the weights belonging to the chunks
selected in a single stage are read from the memory again in the subsequent stage,
and so on.
Fortunately, the total number of times the weights are read from the memory
equals
$\sum_{k = 0}^{ \left \lceil log_p n \right \rceil } \frac{n}{p^k}
\le n \sum_{k = 0}^{\infty} \frac{1}{p^k} = n \frac{p}{p - 1}$,
which is $O(n)$ because the number of threads, $p$, is constant.

\subsubsection{Weighted Reservoir Sampling}
\label{sec:Weighted_Reservoir_Sampling}

The problem of the selection of the next node in the MMAS can also be seen as an
instance of random weighted sampling without replacement, or, more briefly,
weighted reservoir sampling (WRS).
In WRS, one has to select $m$ distinct items randomly out of a population of size
$n$, while the probability of choosing an item is proportional to its
\emph{weight}~\cite{Efraimidis2006}.
In the case of the MMAS, only one item (node) needs to be selected, and the
weights are products of the heuristic information values and the pheromone trails
values (see Eq.~(\ref{eq:prob})) that are stored in the \emph{choice\_info} matrix.

%(fold) alg:wrs
\begin{algorithm}[h]
\SetKwInOut{Input}{Input}\SetKwInOut{Output}{Output}

\Input{ A population $V$ of $n$ weighted items}
\Output{A reservoir (sample) $R$ with the WRS of size $m$}

Insert first $m$ items of $V$ into $R$ \;
\For{ $i \leftarrow 1$ \KwTo $m$ }{
    $k_i \leftarrow u_i^{(1/w_i)}$ where $u_i = \textrm{random}(0,1)$\;
}
\For{$i \leftarrow m+1$ \KwTo $n$}{
    $T \leftarrow $ the smallest key in $R$ \;
    $k_i \leftarrow u_i^{(1/w_i)}$ where $u_i = \textrm{random}(0,1)$\;
    \If{ $k_i > T$ }{
        The item with the minimum key in $R$ is replaced by item $v_i$ \;
    }
}
\caption{Algorithm A-Res for computing WRS~\cite{Efraimidis2006}.}
\label{alg:wrs}
\end{algorithm}
%(end)

Efraimidis and Spirakis proposed an efficient algorithm, named \emph{A-Res}
(Fig.~\ref{alg:wrs}), for computing WRS~\cite{Efraimidis2006}.
The A-Res algorithm assigns each item a key, $u_i ^ {(1/w_i)}$, where $u_i$ is
a uniformly chosen number from the range $(0, 1)$, and then selects $m$ items
with the \emph{largest keys}.
The most important property of the algorithm is that it
selects the sample in \emph{one pass}, i.e., it considers each item and its
weight \emph{once} and, in contrast to the RWM, does not require the summation of
all the weights.
It also worth noting that the relative order in which the items are processed
can be arbitrary, what makes the algorithm easier to parallelize.
Therefore, we propose to adapt this algorithm to implement a parallel
equivalent of the RWM for application in the MMAS.
%(fold) alg:par-wrs
\begin{algorithm}[h]
\SetKw{KwBy}{by}
$t \leftarrow \textrm{threadIdx.x}$ \quad \tcp{CUDA-based thread id}
$p \leftarrow \textrm{blockDim.x}$ \quad \tcp{Number of threads in a block}
$R_t \leftarrow \emptyset $ \quad \tcp{No element was selected}  \label{alg:par-wrs:init1}
$T_t \leftarrow 0$  \quad \tcp{Initial key for the thread $t$} \label{alg:par-wrs:init2}
$l \leftarrow$ tabu.length()

\ForPar(\tcp*[f]{Thread $t$ processes indices: $t, t + p, \ldots, \lfloor \frac{n - t}{p} \rfloor p + t$}){$i \leftarrow t$ \KwTo $l-1$ \KwBy $p$ }{
    \label{alg:par-wrs:for-beg}
     $v \leftarrow $ tabu.get\_candidate(i)
     
     \If(\quad \tcp*[h]{$v$ is an unvisited node}){ $v \ne $ sentinel }{
         $r \leftarrow random(0, 1)$ \\
         $w \leftarrow \left[ \tau_{uv} \right]^\alpha \left[ \eta_{uv} \right ]^\beta$  \label{alg:par-wrs:weight} \\
         $k \leftarrow r^{(1 / w)}$  \label{alg:par-wrs:key}
         
         \If{ $k > T_t$ }{  \label{alg:par-wrs:cmp}
             $T_t \leftarrow k$ \\
             $R_t \leftarrow v$
         }
     }
    \label{alg:par-wrs:for-end}
}
$k \leftarrow \argmax_{i \in \{ 0, 1, \ldots, p-1 \}} T_i$ \quad \tcp{Parallel
    reduction of $(T_0, T_1, \ldots, T_{p-1})$}  
    \label{alg:par-wrs:reduction}
\Return{$R_k$} 
\caption{The WRS-based parallel pseudo-random proportional selection of the next node in the MMAS.}
\label{alg:par-wrs}
\end{algorithm}
%(end)

\begin{figure}[h]
\centering
\includegraphics[]{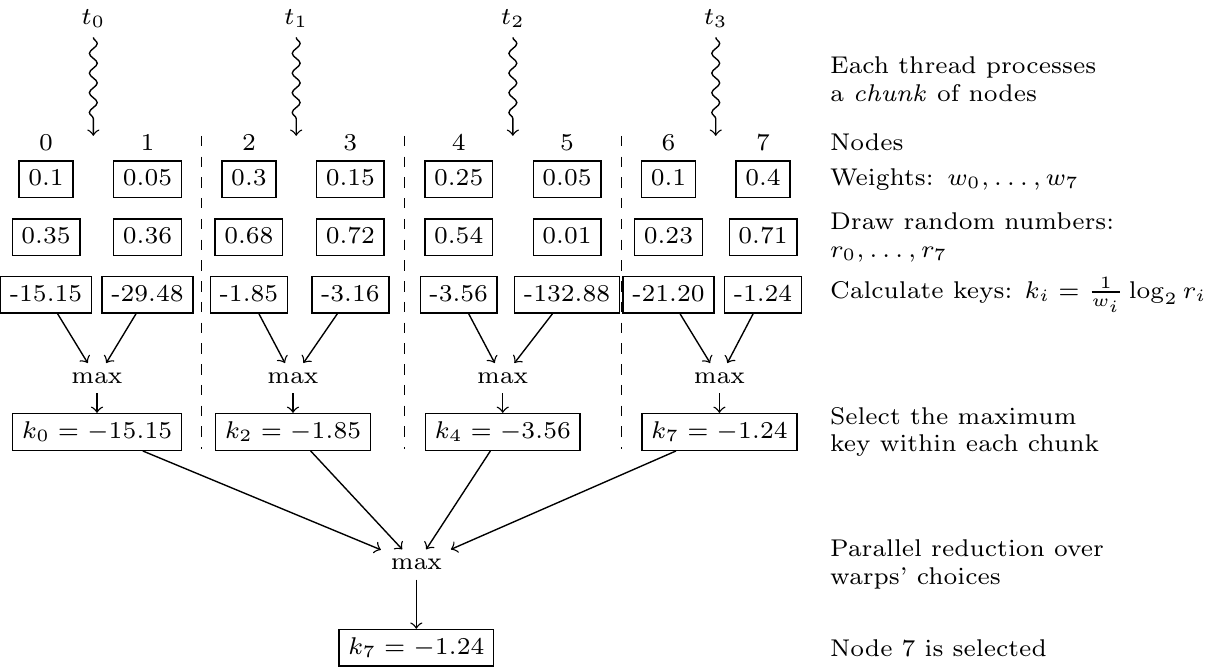}
\caption{
    An example showing the execution of the parallel, WRS-based pseudo-random
    proportional selection of the next node in the MMAS.
}
\label{fig:par-wrs}
\end{figure}

Figure~\ref{alg:par-wrs} presents a pseudocode of the parallel version
of the WRS (a single thread-block is assumed) adapted to perform the 
pseudo-random proportional selection of the next node in the MMAS.
Each thread starts with its own reservoir that has a size of one (as only one
node has to be selected) and the corresponding key (lines
\ref{alg:par-wrs:init1} and \ref{alg:par-wrs:init2}).
Next, the elements of the tabu are processed by $p$ threads in parallel
(loop in lines \ref{alg:par-wrs:for-beg}--\ref{alg:par-wrs:for-end}).
A thread, $t$, processes every $p$-th element and selects its own (locally) maximum
key, $T_t$, and the corresponding element (node) $R_t$.
After this, a parallel reduction of the keys selected by the threads, $(T_0,
T_1, \ldots, T_p)$, is performed and the (final) maximum key is elected (line
\ref{alg:par-wrs:reduction}).
The corresponding element becomes the result of the WRS.
The algorithm selects a node in $O(\frac{l}{p} + \log p)$ time.
Figure~\ref{fig:par-wrs} shows an example of the WRS-based node selection.

%(fold) fig:par-wrs
%(end)

The algorithm has a few potential drawbacks that can negatively affect its
runtime.
Firstly, a single random number is required per element of the tabu, hence a
separate pseudo-random number generator's state needs to be stored for
each thread.
Secondly, the calculation of the keys (line \ref{alg:par-wrs:key} in
Fig.~\ref{alg:par-wrs}) involves costly operations on float numbers: division
and exponentiation.
The former can be alleviated if the reciprocal of each weight (line
\ref{alg:par-wrs:weight}) is calculated in advance.
This is possible because the weights depend on the heuristic information
values that are constant, while the values of the pheromone trails in the MMAS
are updated only once per iteration -- after each ant has completed
construction of its solution.
The latter can be computed using the \texttt{powf()} CUDA function but another
important problem arises -- as the weights are often very small, their
reciprocals become large, and therefore so do their exponents.
This in turn causes problems, as the 32-bit floating point type does
not provide enough precision for the calculations.
Fortunately, it is possible to replace the exponentiation with a logarithm,
i.e., the calculation in line~\ref{alg:par-wrs:key} can be replaced by
$k \leftarrow \frac{1}{w} \log_2 r$, where $r$ is a uniform random number in
the range $(0, 1)$
(Also, the initial $T_t$ value needs to be set carefully as $\lim_{x \to 0^+}
\log_2 x = -\infty$).
In fact, it is now possible to use the fast approximate logarithm calculation
provided by the CUDA \texttt{\_\_log2f()} (intrinsic) function.

\subsection{Candidate lists}
\label{sec:Candidate_lists}

During the solution construction process an ant located at a given node selects
the next node from the set of (yet) unvisited neighboring nodes.
If the set of nodes the ant can choose from is limited to the so called
\emph{candidate list}, the computation time of the MMAS can be greatly reduced,
often without sacrificing quality~\cite{Stutzle2000}.
The candidate list for each node consists of a number, \emph{cl}, of the
closest neighboring nodes. Assuming that $\textit{cl}$ is constant, the time
complexity of the solution construction process for a single ant is reduced
from $O(n^2)$ to $O(n)$.  

In the original (CPU-based) MMAS implementation for solving the TSP,
$\textit{cl} = 20$~\cite{Stutzle2000}.
Even smaller numbers are possible, however it requires a more complex
definition of the \emph{closeness} between nodes, e.g., the
$\alpha$-measure~\cite{Helsgaun2009}.
In the case of the GPU-based computations, setting $\textit{cl}$ to a multiple
of the warp size (32 in case of the modern Nvidia GPUs) seems an obvious
choice~\cite{Dawson2013}.

\subsection{Final details}
\label{sec:Final_details}

\begin{figure}[h]
\centering
\includegraphics[]{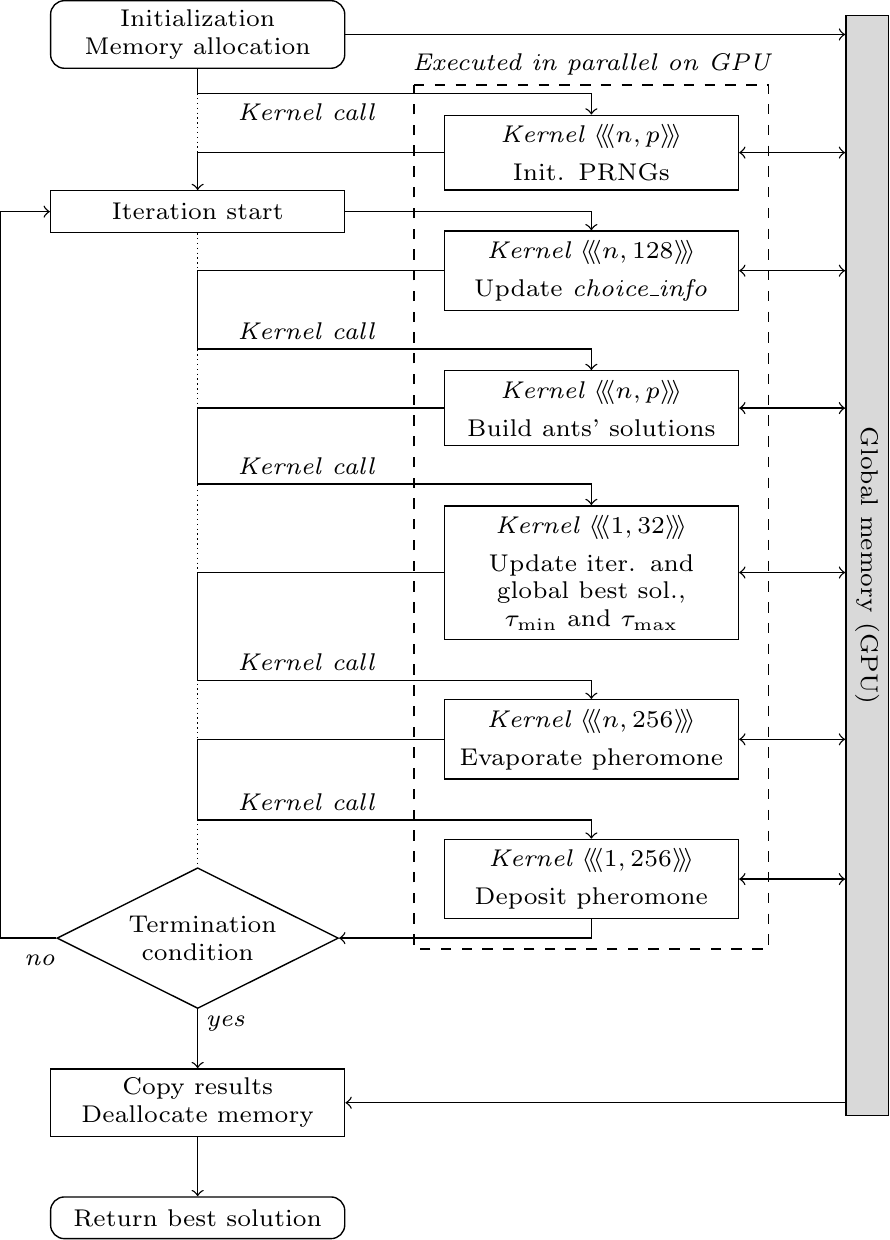}
\caption{
Flowchart of the GPU-based MMAS ($n$ is the size of the problem, $p$ is the
    number of threads per ant).  The pairs of numbers in the angle brackets
    denote the number of thread blocks and the number of threads per block that
    are executing a kernel, respectively.
}
\label{fig:flowchart}
\end{figure}

The final structure of the proposed MMAS implementation is shown in
Fig.~\ref{fig:flowchart}.  The main component of the parallel MMAS is the ants'
solution construction process computed using a single CUDA kernel. We apply the
proven data-parallel approach~\cite{Cecilia2012,Dawson2013,Uchida2012} in which
a single thread-block computes the ant's solution, and the size of each block
is a multiple of the warp size, which is 32  in the Nvidia GPUs, so that all
the processing elements (CUDA cores) within a warp are used efficiently.

%(fold) fig:flowchart
%(end)

After the solutions construction kernel is executed, the iteration best
solution is selected.
If its cost is lower than the cost of the current global best solution, it
becomes the new global best solution and the pheromone trail limits, $\tau_{\rm
min}$ and $\tau_{\rm max}$, are updated.
The new limits are then used by the pheromone evaporation kernel.
This kernel is executed with one thread-block per row of the pheromone memory
matrix, and with 256 threads in every block.
Finally, a pheromone deposition kernel is called.
It is responsible for increasing the values of the pheromone trails that
correspond to the edges of the iteration best solution.
This kernel is executed by a single block of 256 threads.

The presented pheromone evaporation-deposition scheme is simpler than in the
AS, in which the pheromone is updated for every ant and where the changes are
often conflicting, i.e., they concern the same pheromone trails~\cite{Cecilia2012}.
In the MMAS, both the evaporation and deposition of the pheromone can be split
into \emph{independent} and non-conflicting parts that do not require atomic
operations as are used in the GPU-based AS
implementations~\cite{Cecilia2012,Cecilia2018}.

\section{Experimental results}
\label{sec:Experimental_results}

\label{sec:experimental-results}

In this section we present the results of the computational experiments
conducted in order to evaluate the efficiency of the proposed MMAS implementations.
The computations were performed on a set of symmetric TSP instances from the
well-known TSPLIB~\cite{Reinelt1991} repository. The instances were selected so
that the results could be compared to the reports available in the literature.

Following the works by Cecilia et. al~\cite{Cecilia2012,Cecilia2018}, we set the
MMAS parameters as follows:
\begin{itemize}
\item
    the number of ants was equal to the problem size, i.e., $m = n$, where $n$ denotes the size of the TSP instance;
\item
    $\rho = 0.5$ -- the parameter controlling the pheromone evaporation rate;
\item
    $\alpha = 1$ and $\beta = 2$ -- the parameters controlling the influence of the pheromone and the heuristic information on the next node selection probability~(Eq.~\ref{eq:prob});
\item
    the number of iterations equaled 100.
\end{itemize}
The pheromone trail limits, $\tau_{\rm min}$ and $\tau_{\rm max}$, were
calculated following the work of St{\"{u}}tzle and Hoos~\cite{Stutzle2000} with
$p_{\rm best} = 0.01$. The initial values of the limits were set based on the
value of a solution constructed using the nearest neighbor heuristic.  If not
stated otherwise, the presented numbers were averaged over 30 repeated
executions of a given algorithm.  The presented time measurements were obtained
using CUDA provided timers and refer to the \emph{kernels} executing respective
parts of the MMAS.

Almost all of the computation time in the AS and other ACO algorithms is spent
in the solutions construction phase and relatively little on the pheromone
updates~\cite{Cecilia2012}. This proportion is even higher in the MMAS in which
only a single ant deposits pheromone, i.e., the current global or iteration
best, depending on the chosen strategy~\cite{Dawson2013}.  For this reason, in
the computational experiments, we mainly focused on the impact of the tabu
and the implementations of the node selection procedure.

Combining the three tabu implementations presented in Section~\ref{sec:tabu} and
the two node selection procedures in Section~\ref{sec:next.node.selection} results
in a total of six MMAS variants under consideration.
For convenience, they are denoted as MMAS--\emph{node selection
procedure}--\emph{tabu implementation}, where the \emph{node selection procedure}
is either denoted by RWM or WRS; and \emph{tabu implementation} is denoted by one of the
following three: LC, BT and CT tabu implementations.

\subsection{Computing environment}
\label{sec:Computing_environment}

%(fold) tab:gpu-spec
\begin{table}[]
\footnotesize
\centering
\caption{Characteristics of the GPUs used in the computational experiments.}
\label{tab:gpu-spec}
\begin{tabular}{@{}lrr@{}}
\toprule
Characteristic & \multicolumn{1}{l}{Nvidia Tesla P100} & \multicolumn{1}{l}{Nvidia Tesla V100} \\ \midrule
Architecture & Pascal & Volta \\
Multiprocessors (SM) & 56 & 80 \\
Streaming processors (CUDA cores) & 3584 & 5120 \\
FP32 processing power & 9,340 GFLOPS & 14,028 GFLOPS \\
Memory size & 16 GB & 16 GB \\
Memory bandwidth & 732 GB/s & 900 GB/s \\
L2 cache size (per die) & 4,096 KB & 6,144 KB \\
Shared mem. size (per SM) & 64 KB & Up to 96 KB \\
Transistors count & 15 Billion & 21.1 Billion \\
Maximum TDP & 300W & 300W \\ \bottomrule
\end{tabular}
\end{table}
%(end)

The implementation of the proposed algorithms was done in C++ using Nvidia CUDA version 10.
Sources~\footnote{Sources are available at
https://github.com/RSkinderowicz/GPU-based-MMAS} were compiled using GCC v6.3
with a \emph{-O3} switch for the CPU-side code, while the GPU-side
code was compiled with a \emph{-gencode arch=compute\_60,code=sm\_60} switch for the Nvidia Tesla P100 GPU,
and \emph{-gencode arch=compute\_70,code=sm\_70} switch for the Nvidia Tesla V100 GPU.
Table~\ref{tab:gpu-spec} presents the characteristics of the two GPU architectures used.
The computations were conducted on servers running under the Debian 9 Linux OS and
equipped with a 20-core Intel Xeon 6138 (Skylake) CPU clocked at 2 GHz (a single
core was used in the computations).

\subsection{Solution construction phase}
\label{sec:Solution_construction_phase}

The efficient use of the computing power offered by GPUs equipped with
thousands of processing elements (CUDA cores), requires the proper
organization of the computations, i.e., the computations should also be split
into multiple, mostly independent portions~\cite{Kirk2016}.
In the proposed MMAS implementations we adopted the data-parallel approach in
which each ant is assigned a thread-block, and the threads within the
thread-blocks are responsible for computing the solution.
This leaves one crucial decision: the number of threads within a thread-block.
To best adapt to the Nvidia GPUs used in the computations, the size of each
thread block was a multiple of the warp size, that is, 32.
Although the Volta architecture allows the threads within a warp to follow
divergent paths simultaneously~\cite{Jia2018}, setting the number of threads to
a multiple of the warp size simplifies the implementation and allows for
efficient execution on the pre-Volta generations of GPUs, i.e., Pascal.

%(fold) fig:warps-vs-time-pcb1173-p100
\begin{figure}[h]
  \centering
    \includegraphics{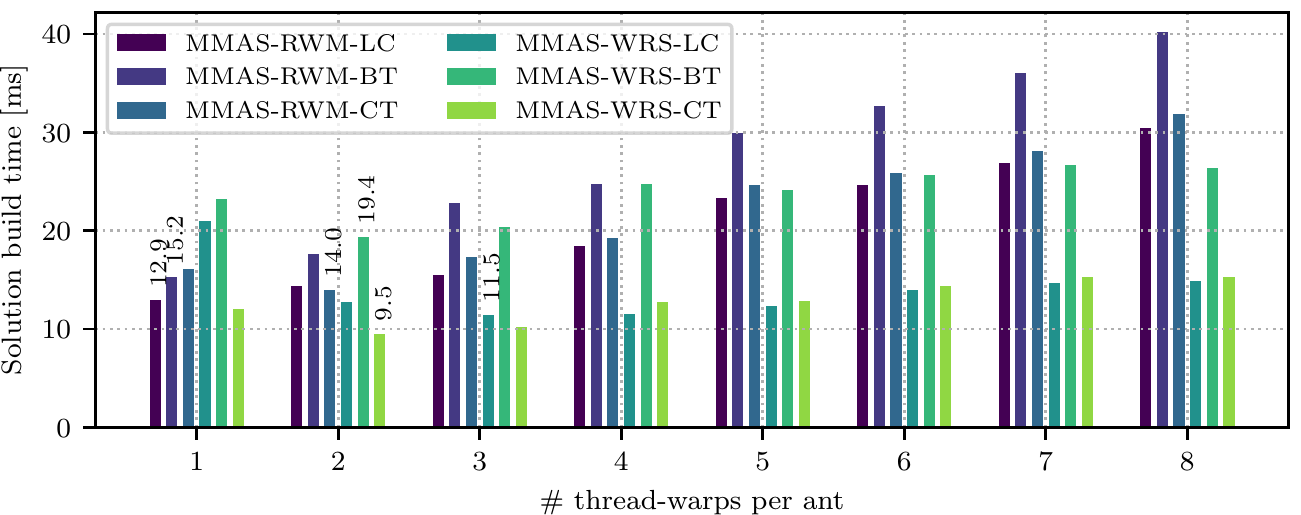}
  \caption{
      The mean time required by the ants to construct solutions in the MMAS
      vs. the number of thread-warps per ant for the \emph{pcb1173} instance 
      ($n$ is the size of the problem). The lowest value for each
      MMAS variant is shown above the respective bar.
      The results are for the Nvidia Pascal P100 GPU.
  }
  \label{fig:warps-vs-time-pcb1173-p100}
\end{figure}

\begin{figure}[h]
  \centering
    \includegraphics{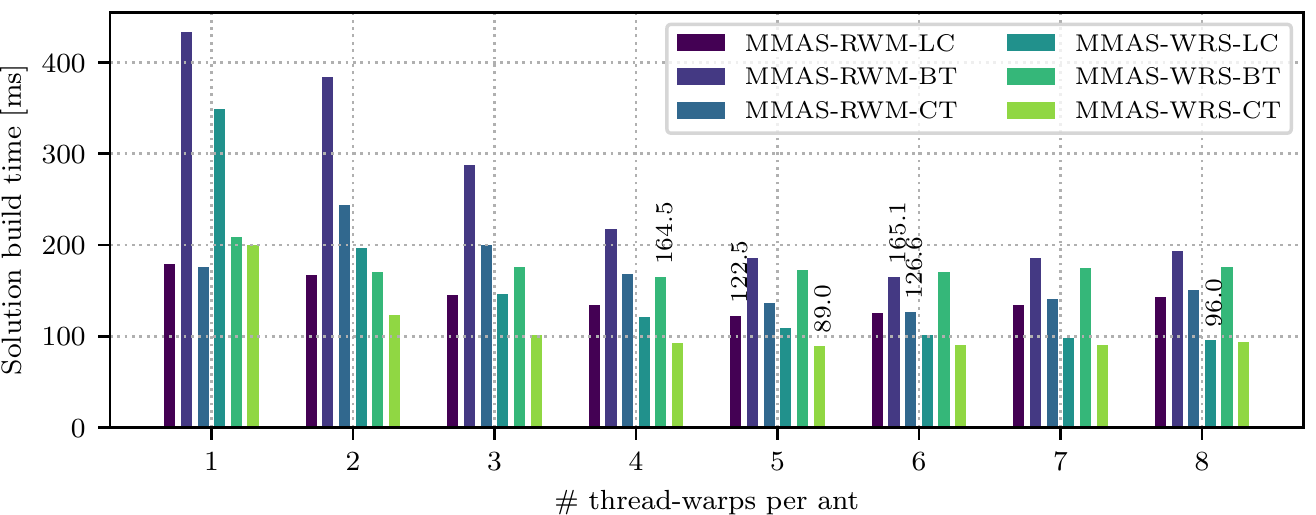}
  \caption{
      The mean time required by the ants to construct solutions in the MMAS
      vs. the number of thread-warps per ant for the \emph{pr2392} instance 
      ($n$ is the size of the problem). The lowest value for each
      MMAS variant is shown above the respective bar.
      The results are for the \emph{Nvidia P100 GPU}.
  }
  \label{fig:warps-vs-time-pr2392-p100}
\end{figure}
%(end)

%(fold) fig:p100-occupancy-pr2392
\begin{figure}
  \centering
    \includegraphics{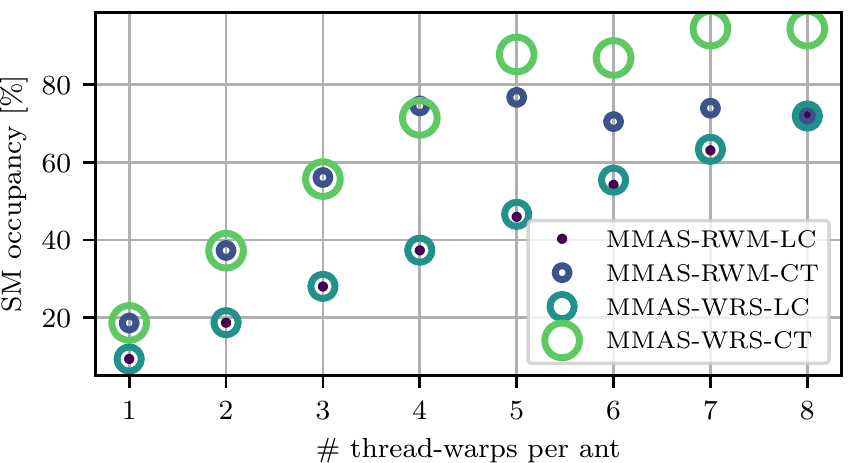}
  \caption{
  SM occupancy vs. the number of thread warps for the \emph{pr2392} instance and  
  the \emph{Nvidia P100 GPU}.
  }
  \label{fig:p100-occupancy-pr2392}
\end{figure}
%(end)

Figures~\ref{fig:warps-vs-time-pcb1173-p100}
and~\ref{fig:warps-vs-time-pr2392-p100} show the mean solution construction
time for the proposed MMAS variants that executed on the Nvidia P100 GPU vs the
number of threads per ant obtained for the \emph{pcb1173} and \emph{pr2392} TSP
instances, respectively.
A few observations can be made here.
First, the MMAS variants differ significantly in speed depending both on the
node selection and the tabu implementation.
The fastest is the MMAS-WRS-CT variant followed by
the MMAS-WRS-LC.
The slowest are the variants with the BT, which can be explained by the fact
that the BT does not provide direct access to the list of nodes to visit, but
checks the status (and the corresponding bit) of each node repeatedly, hence
significantly increasing the number of computations.

Second, the parallel WRS-based node selection procedure is faster and \emph{scales}
better than the parallel RWM implementation.
Even though the WRS method involves costly computations (i.e., the random
number generation and logarithm calculation), it reduces the number of data
exchanges and synchronizations between the threads.

Third, the results indicate that the number of threads should be adjusted to the
size of the problem.
Specifically, the bigger the problem is, the higher the number of threads per
ant are required.
This can be observed when comparing the results for
the 1,173-city instance (Fig.~\ref{fig:warps-vs-time-pcb1173-p100}) to the
results for the 2,392-city instance (Fig.~\ref{fig:warps-vs-time-pr2392-p100}).
For the former, the fastest computations were observed for 32 and 64 threads (1
and 2 warps, respectively) per ant; while for the latter, the fastest execution
was observed when the number of threads per ant was at least 128 (4 warps).
Generally, the efficient use of the GPU's computing power is related to its
\emph{occupancy}: understood as the ratio of the number of \emph{active}
thread warps to the maximum number of warps per SM~\cite{Kirk2016}.
In most cases, this ratio should be high.
Often, having more active threads than processing
elements (CUDA cores) is beneficial as it allows latencies generated by
global memory accesses and synchronization operations to be hidden.
The number of thread blocks (each consisting of thread warps) that can execute
simultaneously on the same SM is limited by the size of the shared memory and
the number of registers used.
By increasing the number of ants, we are increasing the number of thread
blocks.
The LC tabu takes twice as much shared memory as the CT.
Thus the latter allows more thread blocks per SM to be executed at the same time.
Another, complementary, solution that increases the occupancy is to increase the
number of threads per ant.
A comparison of occupancy for the proposed MMAS
variants with LC and CT tabu implementations solving the \emph{pr2392} TSP
instance, is shown in Fig.~\ref{fig:p100-occupancy-pr2392}.
As can be seen, the CT allows for a higher occupancy, with the MMAS-WRS-CT
achieving values exceeding  90\% for 7 or 8 thread warps per ant; while for the
MMAS with the LC tabu the occupancy does not exceed 80\%.

%(fold)fig:warps-vs-time-pr2392-v100
\begin{figure}
  \centering
    \includegraphics{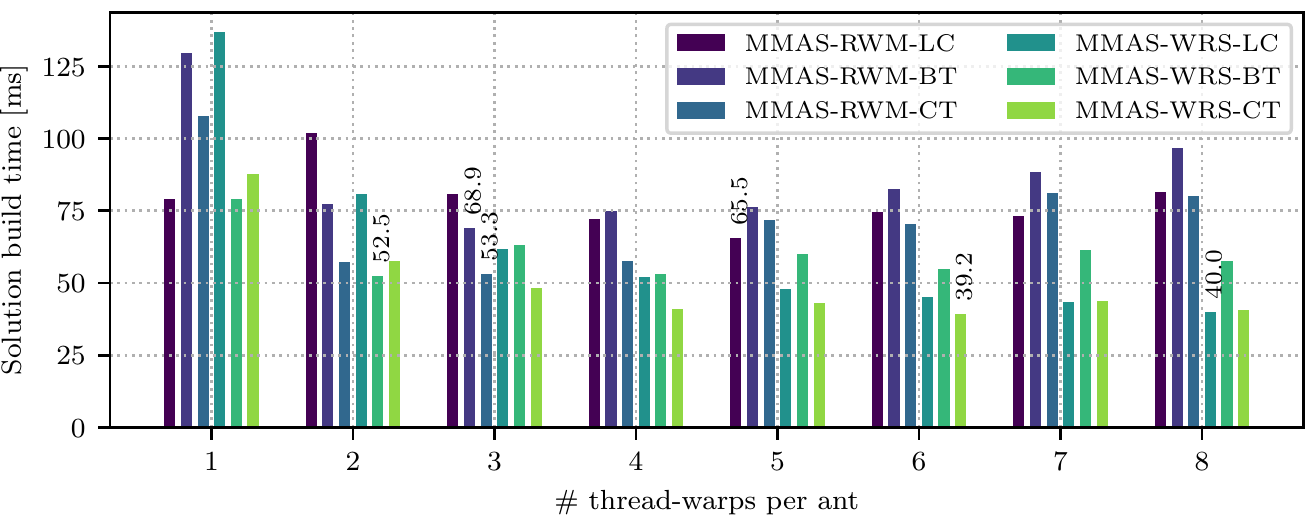}
  \caption{
      The mean time required for the ants to construct solutions in the MMAS
      vs. the number of thread-warps per ant for the \emph{pr2392} instance 
      ($n$ is the size of the problem). The lowest value for each
      MMAS variant is shown above the respective bar.
      The results are for the \emph{Nvidia V100 GPU}.
  }
  \label{fig:warps-vs-time-pr2392-v100}
\end{figure}
%(end)

Executing the same MMAS variants on the \emph{Nvidia Volta V100} GPU reveals a
significant advantage in computing power over the Pascal architecture.
This can be seen in Fig.~\ref{fig:warps-vs-time-pr2392-v100}, which shows a
comparison of the execution times of the solution construction phase for the
\emph{pr2392} instance.
The fastest variant is again the MMAS-WRS-CT taking about 39.2~ms to
complete, vs. 89~ms for the previous generation of GPU.
This difference can be somewhat surprising as it cannot be explained simply by
the higher processing power, which increased by only about 50\%, i.e., 9,340 to
14,028 GFLOP/s.
Rather, the
advantage can be attributed to \emph{a combination} of multiple factors
including an increased number of the processing elements (cores) (see
Tab.~\ref{tab:gpu-spec}), a larger shared memory per SM, a larger L2 cache, a
higher global memory bandwidth and an improved L1 cache replacement policy,
among others~\cite{Jia2018}.

%(fold) tab:mmas-cmp-p100
\begin{table}[]
\footnotesize
\centering
\caption{
A comparison of the mean solution construction phase times (in ms) for the
proposed MMAS variants executed on the \emph{Nvidia P100} GPU.
The numbers in parentheses denote the number of thread warps per
block (ant). The smallest time for each instance was marked in bold.
}
\label{tab:mmas-cmp-p100}
\begin{tabular}{@{}lrrrrrrrr@{}}
\toprule
\multirow{2}{*}{Algorithm} & \multicolumn{8}{c}{Instance} \\
 & \multicolumn{1}{c}{\textit{d198}} & \multicolumn{1}{c}{\textit{pcb442}} & \multicolumn{1}{c}{\textit{rat783}} & \multicolumn{1}{c}{\textit{pr1002}} & \multicolumn{1}{c}{\textit{pcb1173}} & \multicolumn{1}{c}{\textit{rl1889}} & \multicolumn{1}{c}{\textit{pr2392}} & \multicolumn{1}{c}{\textit{fl3795}} \\ \cmidrule(r){1-1} \cmidrule(l){2-9} 
\#1: AS~(CPU)~\cite{Zhou2018} & 0.87 & 5.66 & 25.25 & 47.17 & NA & NA & 822.54 & NA \\
\#2: AS~(GPU)~\cite{Cecilia2018} & 0.40 & 1.59 & 6.81 & 15.54 & 23.53 & 78.87 & 185.13 & 726.25 \\ 
\cmidrule(r){1-1} \cmidrule(l){2-9}
MMAS-RWM-LC & 0.33 (1) & 1.05 (1) & \textbf{2.75 (1)} & 7.77 (1) & 12.92 (1) & 62.26 (4) & 122.46 (5) & 482.60 (8) \\
MMAS-RWM-BT & 0.76 (3) & 1.37 (1) & 3.67 (1) & 8.59 (1) & 15.25 (1) & 83.50 (5) & 165.11 (6) & 656.51 (10) \\
MMAS-RWM-CT & 0.47 (1) & 1.15 (1) & 2.80 (1) & 6.45 (1) & 14.00 (2) & 65.05 (4) & 126.63 (6) & 521.57 (10) \\
MMAS-WRS-LC & \textbf{0.30 (2)} & \textbf{0.92 (2)} & 3.06 (2) & 7.68 (3) & 11.46 (3) & 48.59 (6) & 95.96 (8) & 383.67 (10) \\
MMAS-WRS-BT & 0.42 (4) & 1.58 (2) & 6.33 (2) & 12.17 (2) & 19.37 (2) & 94.87 (5) & 164.54 (4) & 652.62 (6) \\
MMAS-WRS-CT & 0.31 (3) & 0.95 (2) & 3.18 (2) & \textbf{6.09 (2)} & \textbf{9.54 (2)} & \textbf{46.02 (5)} & \textbf{88.99 (5)} & \textbf{354.43 (10)} \\ \cmidrule(r){1-1} \cmidrule(l){2-9}
Speedup vs. \#1 & 2.90x & 6.14x & 9.19x & 7.74x & - & - & 9.24x & - \\
Speedup vs. \#2 & 1.33x & 1.73x & 2.48x & 2.55x & 2.47x & 1.71x & 2.08x & 2.05x \\ \bottomrule
\end{tabular}
\end{table}
%(end)

Table~\ref{tab:mmas-cmp-p100} presents a summary of the mean times needed to
execute the MMAS solution construction kernel for the proposed MMAS variants
for the Nvidia Tesla P100 GPU.  Generally, the MMAS with the WRS-based node
selection implementation was faster in all cases but one, while the memory
savings resulting from the CT had an impact only for sufficiently large
instances, i.e., those with at least 1,000 nodes.

For reference point, we have also provided the results from the two recent works by
Zhou et al.~\cite{Zhou2018} and Cecilia et al.~\cite{Cecilia2018}.
Although in both cases the authors proposed parallel implementations of the AS,
the main differences between the AS and the MMAS relate to the pheromone
update, while the solution construction phase is analogous, assuming that no
candidate lists are used~\cite{Stutzle2000}.
Specifically, the numbers from Zhou et
al.~\cite{Zhou2018} refer to a parallel, multicore-SIMD CPU based AS
implementation named VETPAM-CPU-AS, executed on a six-core Intel i7-5820
(Haswell) CPU.
The numbers from Cecilia et al.~\cite{Cecilia2018} are
the execution times of the solution construction kernel for the CUDA-based AS
implementation running on an Nvidia GTX TITAN X GPU with 3,072 CUDA cores and
6.69 GFLOP/s of computing power (Maxwell architecture).
In comparison, the slower of our GPUs, the Nvidia P100, has about 52\% more
computing power (single precision), 25\% larger L2 cache and more than double
the global memory bandwidth (732 GB/s vs.~336 GB/s).
The results show that the proposed
MMAS-WRS-CT is up to 2.55x faster, which confirms that our implementation
utilizes the additional computing resources efficiently.
In comparison, the highly optimized VETPAM-CPU-AS is up to 9.24x slower, though
still, being very fast if we account for the relatively low computing power of
the CPU used for the computations (316 GFLOP/s).

%(fold) tab:mmas-cmp-v100
\begin{table}[]
\footnotesize
\centering
\caption{
    The comparison of duration (in ms) of the mean solution construction phase
    for the proposed MMAS variants executed on the \emph{Nvidia V100} GPU.  The
    numbers in parentheses denote the number of thread warps per block (ant).
    The smallest time for each instance is marked in bold.
}
\label{tab:mmas-cmp-v100}
\begin{tabular}{@{}lrrrrrrrr@{}}
\toprule
\multirow{2}{*}{Algorithm} & \multicolumn{8}{c}{Instance} \\
 & \multicolumn{1}{c}{\textit{d198}} & \multicolumn{1}{c}{\textit{pcb442}} & \multicolumn{1}{c}{\textit{rat783}} & \multicolumn{1}{c}{\textit{pr1002}} & \multicolumn{1}{c}{\textit{pcb1173}} & \multicolumn{1}{c}{\textit{rl1889}} & \multicolumn{1}{c}{\textit{pr2392}} & \multicolumn{1}{c}{\textit{fl3795}} \\ \cmidrule(r){1-1} \cmidrule(l){2-9}
\#1: AS~(CPU)~\cite{Zhou2018} & 0.87 & 5.66 & 25.25 & 47.17 & NA & NA & 822.54 & NA \\
\#2: AS~(GPU)~\cite{Cecilia2018} & 0.40 & 1.59 & 6.81 & 15.54 & 23.53 & 78.87 & 185.13 & 726.25 \\
\cmidrule(r){1-1} \cmidrule(l){2-9}
MMAS-RWM-LC & 0.28 (6) & 0.77 (1) & 1.87 (1) & 3.08 (1) & 5.21 (1) & 32.75 (4) & 65.50 (5) & 285.72 (6) \\
MMAS-RWM-BT & 0.42 (1) & 1.22 (1) & 2.84 (2) & 5.04 (2) & 7.66 (1) & 31.20 (2) & 68.90 (3) & 301.75 (6) \\
MMAS-RWM-CT & 0.29 (6) & 0.81 (1) & 1.85 (1) & 3.23 (1) & 5.55 (1) & 24.06 (2) & 53.26 (3) & 250.62 (6) \\
MMAS-WRS-LC & \textbf{0.18 (3)} & \textbf{0.50 (5)} & \textbf{1.26 (4)} & \textbf{2.16 (3)} & 3.36 (3) & 20.39 (5) & 40.03 (8) & 196.00 (10) \\
MMAS-WRS-BT & 0.23 (4) & 0.70 (5) & 2.19 (4) & 3.85 (4) & 5.97 (4) & 28.36 (5) & 52.54 (2) & 236.14 (8) \\
MMAS-WRS-CT & 0.19 (6) & 0.52 (5) & 1.30 (4) & 2.20 (3) & \textbf{3.31 (3)} & \textbf{19.12 (5)} & \textbf{39.24 (6)} & \textbf{193.05 (10)} \\ \cmidrule(r){1-1} \cmidrule(l){2-9}
Speedup vs. \#1 & 4.77x & 11.42x & 20.07x & 21.79x & - & - & 20.96x & - \\
Speedup vs. \#2 & 2.19x & 3.21x & 5.41x & 7.18x & 7.10x & 4.12x & 4.72x & 3.76x \\ \bottomrule
\end{tabular}
\end{table}
%(end)

\begin{figure}
\centering
\begin{subfigure}{.5\textwidth}
  \centering
  \includegraphics[width=\linewidth]{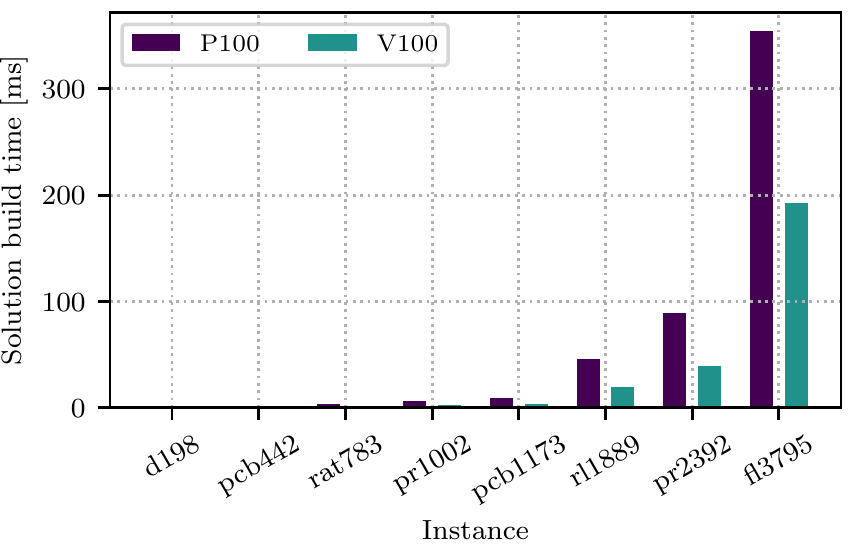}
\end{subfigure}%
\begin{subfigure}{.5\textwidth}
  \centering
  \includegraphics[width=\linewidth]{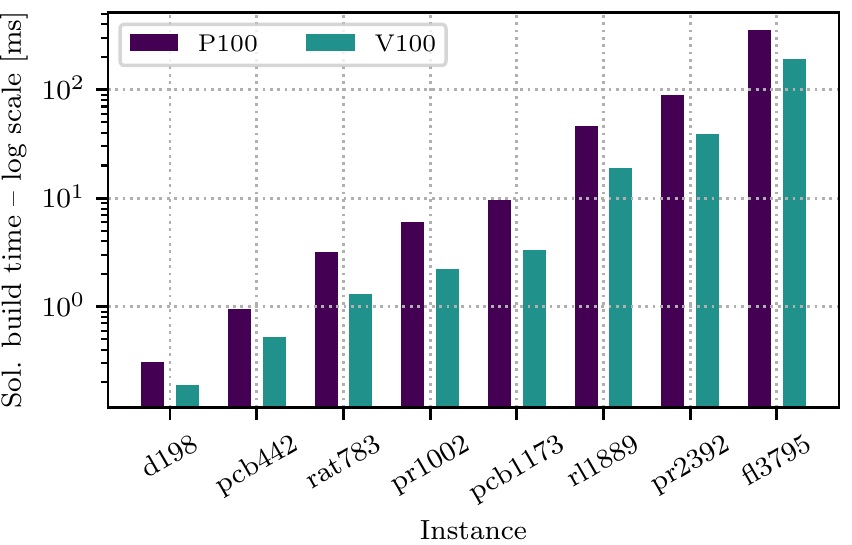}
\end{subfigure}
\caption{
A comparison of the solution construction times for
the TSP instances under consideration, obtained for the P100 and V100 GPUs.
}
\label{fig:gpu-cmp}
\end{figure}

Table~\ref{tab:mmas-cmp-v100} shows the results for the Nvidia Volta GPU
architecture.
In a similar way to the results for the Pascal architecture, the MMASs with the
WRS-based implementation were faster than their counterparts with the RWM-based
implementation.
Specifically, the MMAS-WRS-CT was about 2.88x faster when solving the
\emph{pcb1173} instance, but for the larger instances, the speedup dropped to
only 2x.
This difference can be explained (partially) by the larger
L2 cache present in the V100 GPU, 6MB vs. 4MB in the P100 GPU.
Storing the \emph{choice\_info} data for the \emph{pcb1173} instance takes
up about 5.25MB which fits entirely into the larger cache of the V100 GPU, but
only partially into the L2 of the P100 GPU.
Generally, the newer GPU offers advantages over the older architecture in all
cases, but the differences are larger for medium and large instances, as can be
seen in Figure~\ref{fig:gpu-cmp}.

%(fold) fig:gpu-cmp
%(end)

Compared with the times reported by Cecilia et al.~\cite{Cecilia2018}, the
proposed MMAS implementation is up to 7.18x faster.
At the same time, the advantage over the VETPAM-CPU-AS exceeds
20x for the following instances: \emph{rat783, pr1002} and \emph{pr2392}.

\subsection{Candidate Lists}
\label{sec:Candidate_Lists}

Using candidate lists can have a significant impact on the MMAS convergence
and also on the execution speed~\cite{Stutzle2000}.
In our experiments, the length of the candidate lists was set to 32, following
Dawson and Stewart~\cite{Dawson2013} who proposed a GPU-based AS using the LC
tabu implementation.
This value results in a single thread warp per ant.
It is
worth noting that the V100 GPU has 5,120 CUDA cores, thus we need at least 160
ants to utilize the available computing power.
In fact, this number should be even higher to allow for effective memory
latency hiding through context switching between the active thread
warps~\cite{Kirk2016}.

%(fold) tab:mmas-cl-p100
\begin{table}[]
\footnotesize
\centering
\caption{
The mean solution construction phase duration (in ms) for the
proposed MMAS variants with a \emph{candidate list} length of 32,
executed on the \emph{Nvidia P100 (Pascal)} GPU.
}
\label{tab:mmas-cl-p100}
\begin{tabular}{lrrrrrrrr}
\hline
\multirow{2}{*}{Algorithm} & \multicolumn{8}{c}{Instance} \\
 & \multicolumn{1}{l}{\textit{d198}} & \multicolumn{1}{l}{\textit{pcb442}} & \multicolumn{1}{l}{\textit{rat783}} & \multicolumn{1}{l}{\textit{pr1002}} & \multicolumn{1}{l}{\textit{pcb1173}} & \multicolumn{1}{l}{\textit{rl1889}} & \multicolumn{1}{l}{\textit{pr2392}} & \multicolumn{1}{c}{\textit{fl3795}} \\ 
\cmidrule(r){1-1} \cmidrule(l){2-9}
AS~\cite{Dawson2013} & 0.77 & 3.67 & 12.13 & 19.76 & - & - & 131.85 & - \\
\cmidrule(r){1-1} \cmidrule(l){2-9}
MMAS-RWM-LC & 0.34 & 0.64 & 1.11 & 2.34 & 2.77 & 10.09 & 19.84 & 88.19 \\
MMAS-RWM-BT & 0.35 & 0.67 & 1.20 & 1.70 & 2.09 & \textbf{6.69} & \textbf{9.33} & 28.78 \\
MMAS-RWM-CT & 0.35 & 0.67 & 1.19 & 1.63 & \textbf{2.01} & 7.34 & 12.04 & 52.32 \\
MMAS-WRS-LC & \textbf{0.33} & \textbf{0.62} & \textbf{1.09} & 2.25 & 2.67 & 9.70 & 18.91 & 85.71 \\
MMAS-WRS-BT & 0.34 & 0.64 & 1.17 & 1.69 & 2.13 & 6.75 & 9.44 & \textbf{28.75} \\
MMAS-WRS-CT & 0.35 & 0.65 & 1.16 & \textbf{1.62} & 2.07 & 7.23 & 11.77 & 51.08 \\
\cmidrule(r){1-1} \cmidrule(l){2-9}
Speedup vs. AS~\cite{Dawson2013} & 2.32x & 5.91x & 11.18x & 12.18x & - & - & 14.12x & - \\ \hline
\end{tabular}
\end{table}
%(end)

Table~\ref{tab:mmas-cl-p100} presents the mean duration of the solution
construction phase for the proposed MMAS implementations.
Generally, for the smallest instances, the differences between the algorithms
are also small.
Only when
starting with the \emph{pr1002} instance do the differences increase with
performance, mainly depending on the shared memory's efficiency for the
given tabu implementation, with the BT offering the best performance, and the
LC the worst (Tab.~\ref{tab:tabu-cmp}).
For example, for the largest instance,
\emph{fl3795}, and the MMAS with the WRS node selection implementation; the BT
results in a 1.78x and 2.98x speedup over the CT and LC tabu implementations,
respectively.
The implementation is also up to 14.12x faster than the GPU-based
AS by Dawson and Stewart~\cite{Dawson2013} who used the Nvidia GTX 580 GPU
(Fermi architecture) with 1,581 GFLOP/s computing power and 192 GB/s global
memory bandwidth. 
As the source code is not available, it is difficult to
repeat the computations on the same GPU. However, the observed speedup suggests
that our implementation efficiently utilizes the newer GPU's computing
capacity.

%(fold) tab:mmas-cl-v100
\begin{table}[]
\footnotesize
\centering
\caption{
The mean solution construction phase duration (in ms) for the
proposed MMAS variants with a \emph{candidate list} length of 32,
executed on the \emph{Nvidia V100 (Volta)} GPU.
}
\label{tab:mmas-cl-v100}
\begin{tabular}{@{}lrrrrrrrr@{}}
\toprule
\multirow{2}{*}{Algorithm} & \multicolumn{8}{c}{Instance} \\
 & \multicolumn{1}{c}{\textit{d198}} & \multicolumn{1}{c}{\textit{pcb442}} & \multicolumn{1}{c}{\textit{rat783}} & \multicolumn{1}{c}{\textit{pr1002}} & \multicolumn{1}{c}{\textit{pcb1173}} & \multicolumn{1}{c}{\textit{rl1889}} & \multicolumn{1}{c}{\textit{pr2392}} & \multicolumn{1}{c}{\textit{fl3795}} \\ 
\cmidrule(r){1-1} \cmidrule(l){2-9}
AS~\cite{Dawson2013} & 0.77 & 3.67 & 12.13 & 19.76 & - & - & 131.85 & - \\ 
\cmidrule(r){1-1} \cmidrule(l){2-9}
MMAS-RWM-LC & \textbf{0.19} & 0.40 & 0.72 & 0.95 & 1.13 & 3.54 & 8.18 & 32.16 \\
MMAS-RWM-BT & 0.20 & 0.37 & 0.71 & 0.97 & 1.16 & 2.49 & 5.18 & 12.83 \\
MMAS-RWM-CT & 0.20 & 0.41 & 0.74 & 1.00 & 1.20 & 2.32 & 5.04 & 19.32 \\
MMAS-WRS-LC & \textbf{0.19} & 0.38 & 0.69 & \textbf{0.92} & \textbf{1.08} & 3.40 & 7.74 & 30.77 \\
MMAS-WRS-BT & \textbf{0.19} & \textbf{0.36} & \textbf{0.68} & 0.93 & 1.12 & 2.44 & \textbf{3.26} & \textbf{12.65} \\
MMAS-WRS-CT & \textbf{0.19} & 0.40 & 0.71 & 0.97 & 1.15 & \textbf{2.26} & 4.89 & 18.66 \\
\cmidrule(r){1-1} \cmidrule(l){2-9}
Speedup vs. AS~\cite{Dawson2013} & 4.14x & 10.26x & 17.84x & 21.51x & - & - & 40.39x & - \\ \bottomrule
\end{tabular}
\end{table}
%(end)

Repeating the computations on the newer, Volta V100, GPU results in speedups up
to 2.86x for the \emph{pr2392} instance compared to the Pascal P100 GPU.
However, for the largest instance, \emph{fl3795}, the speedup falls to 2.27x suggesting
that the larger computing capacity of the newer GPU is not able to compensate
for the significantly increased number of high-latency memory accesses associated
with solving the bigger TSP instance.
Compared to the GPU-based AS by Dawson and Stewart~\cite{Dawson2013}, the
proposed MMAS implementations are up to 40.39x faster.
Considering the impact of the node selection procedure and the tabu
implementations, the latter is more important, especially for the largest TSP
instances.

\subsection{Varying the number of ants}
\label{sec:Varying_the_number_of_ants}

%(fold) fig:var-number-of-ants
\begin{figure}
\centering
\begin{subfigure}{.5\textwidth}
  \centering
  \includegraphics[width=\linewidth]{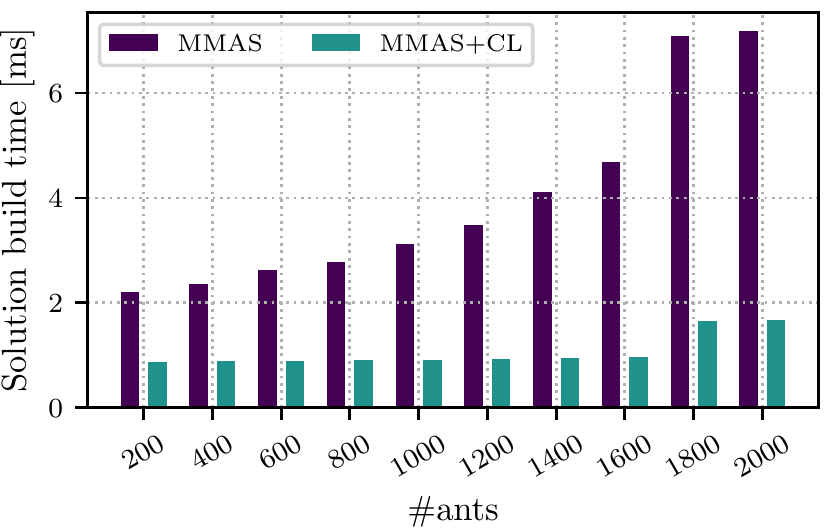}
\end{subfigure}%
\begin{subfigure}{.5\textwidth}
  \centering
  \includegraphics[width=\linewidth]{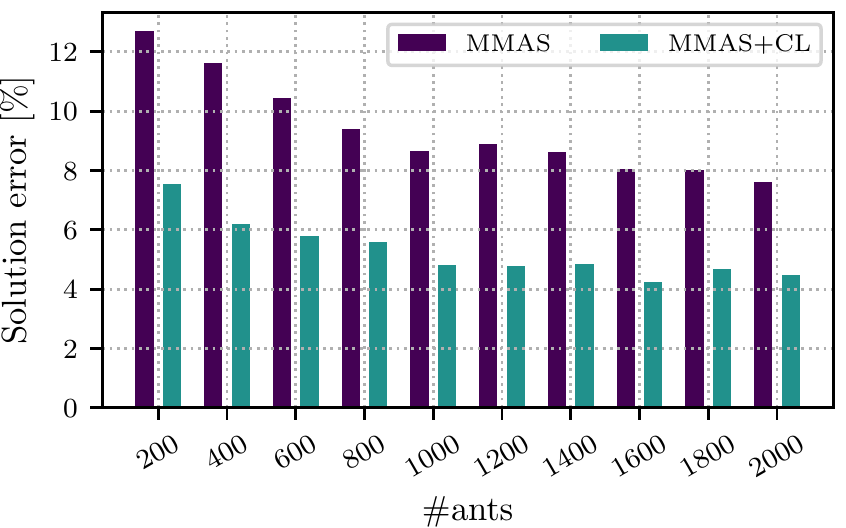}
\end{subfigure}
\caption{
    The influence of the number of ants in the MMAS-RWM-CL on the mean solution
    build time. The results are mean values obtained over 30 runs of the
    algorithm solving the \emph{pr1002} instance on the Nvidia V100 GPU.
}
\label{fig:var-number-of-ants}
\end{figure}
%(end)

In the experiments presented above, the number of ants was equal to the size of
the instance to enable a comparison with the existing results from the
literature~\cite{Cecilia2018,Dawson2013,Zhou2017}.  However nothing prevents
the number of ants being set to a different value.
Figure~\ref{fig:var-number-of-ants} shows how the number of ants affects the
runtime and the quality of the solutions generated by the MMAS-RWM-LC (with and
without the CL) solving the \emph{pr1002} instance on the Nvidia V100 GPU. The
number of iterations was set to 1,000 and the $\rho$ parameter governing the
pheromone evaporation speed was set to $0.9$.

As can be seen, the computation time rises steadily with the increasing number
of ants but not as fast as it might be expected based on the numbers alone.  This can
be explained by the fact that the efficient utilization of the 5,120 computing
cores of the Nvidia Tesla V100 GPU (see Tab.~\ref{tab:gpu-spec}) requires
running a large number of calculations in parallel, i.e., the ants constructing
the solutions. On the other hand, each kernel executing on the GPU requires a
number of the SM's registers and a portion of the shared memory to run.
Combined with other architecture-related restrictions, the number of thread
warps resulting in the best occupancy of the GPU's SMs can be calculated.
For example, the \texttt{cudaOccupancyMaxActiveBlocksPerMultiprocessor}
function offered by the CUDA framework reports that the best occupancy for the
solution construction kernel of the MMAS-RWM-LC (assuming 1 warp per thread
block) can be achieved for 1,680 ants.  This is consistent with the significant
increase in the execution time as observed when going from 1,600 to 1,800 ants.

Focusing on the quality of the solutions, we can observe a positive effect of
increasing the number of ants. However, the benefit is much more visible when
the number of ants goes from 200 to 1,000, than when it goes from 1,000 to 2,000.
In other words, assuming that the number of ants is high enough, it could be
more reasonable to increase the number of iterations than to further increase
the number of ants.

\subsection{Pheromone Update and Remaining Operations}
\label{sec:Pheromone_Update_and_Remaining_Operations}

The solution construction process is definitely the most time consuming part of the MMAS.
The remaining operations, including the pheromone evaporation and deposition,
take significantly less time.
Table~\ref{tab:remaining-times} shows the times needed to complete all the
kernels except for the solution construction time for the MMAS-RWM-LC
algorithm, as executed on the Pascal and Volta GPUs.
The times for the remaining MMAS variants are very close which stems from the
fact that most of the differences between the algorithms affect only the solution
construction process. 
As can be seen, the times for the newer-generation Volta GPU are close to two
times slower than for the older Pascal GPU. 
If we compare the numbers to the solution construction kernel times
(Tab.~\ref{tab:mmas-cmp-p100} and Tab.~\ref{tab:mmas-cmp-v100}), it can be seen
that they are significantly lower and the relative differences increase with
the size of the problem.
For example, for the \emph{pcb442} instance the solution construction kernel
time of the MMAS-WRS-CT executed on the P100 GPU was close to $7.92x$ greater
than the total time needed by all the remaining kernels (0.95 ms vs 0.12 ms).
For the largest instance, \emph{fl3795}, the ratio was close to 246.13x, meaning
that the solution construction process accounted for over 99.5\% of the
computation time.
The ratio remains high ($\sim 35.47x$) even if the construction process is sped
up by using the candidate lists.

It is be possible that a faster execution of the remaining kernels could be
observed if some of them were joined (fused) into one; for example, the
pheromone evaporation and deposition kernels.
However, as the above analysis has shown, it would have little impact on the
overall MMAS execution time while, at the same time, making the implementation
more complicated.

%(fold) tab:remaining-times
\begin{table}[]
\footnotesize
\centering
\caption{
A comparison of the mean times (in ms) necessary to complete the execution of
    all kernels except the solution construction for the MMAS-RWM-LC, with a
    candidate list of 32, executed on the P100 and V100 GPUs.
}
\label{tab:remaining-times}
\begin{tabular}{@{}lcrrrrrrr@{}}
\toprule
\multirow{2}{*}{GPU} & \multicolumn{8}{c}{Instance} \\
 & \textit{d198} & \multicolumn{1}{c}{\textit{pcb442}} & \multicolumn{1}{c}{\textit{rat783}} & \multicolumn{1}{c}{\textit{pr1002}} & \multicolumn{1}{c}{\textit{pcb1173}} & \multicolumn{1}{c}{\textit{rl1889}} & \multicolumn{1}{c}{\textit{pr2392}} & \multicolumn{1}{c}{\textit{fl3795}} \\ \midrule
P100 (Pascal) & \multicolumn{1}{r}{0.08} & 0.12 & 0.21 & 0.28 & 0.35 & 0.61 & 0.81 & 1.44 \\
V100 (Volta) & \multicolumn{1}{r}{0.05} & 0.07 & 0.12 & 0.15 & 0.18 & 0.34 & 0.44 & 0.89 \\
\bottomrule
\end{tabular}
\end{table}
%(end)

\subsection{Solution Quality}
\label{sec:Solution_Quality}

Our work focused on an efficient parallel GPU-based implementation of the
MMAS, hence no quality-related modifications of the MMAS were investigated.
In fact, to confirm that the proposed MMAS variants differ only in terms of
the computation speed, the samples of the final solutions' quality obtained
using the proposed MMAS implementations were checked for statistically significant
differences based on the non-parametric Friedman test~\cite{Hollander2013}.

%(fold) tab:stat-test
\begin{table}[]
\footnotesize
\centering
\caption{
The results of the non-parametric Friedman test with
the null hypothesis $H_0$ stating that the six proposed MMAS
variants produced solutions with the same median quality.
The results are provided for each TSP instance separately.
Assumed significance level of $\alpha = 0.05$.
}
\label{tab:stat-test}
\begin{tabular}{@{}lrrrrrrrr@{}}
\toprule
Instance & \multicolumn{1}{l}{\textit{d198}} & \multicolumn{1}{l}{\textit{pcb442}} & \multicolumn{1}{l}{\textit{rat783}} & \multicolumn{1}{l}{\textit{pr1002}} & \multicolumn{1}{l}{\textit{pcb1173}} & \multicolumn{1}{l}{\textit{rl1889}} & \multicolumn{1}{l}{\textit{pr2392}} & \multicolumn{1}{l}{\textit{fl3795}} \\ \midrule
Test statistic & 2.13 & 2.98 & 3.72 & 8.76 & 3.32 & 6.14 & 6.41 & 4.34 \\
$p$-value & 0.83 & 0.70 & 0.59 & 0.12 & 0.65 & 0.29 & 0.27 & 0.50 \\
$H_0$ rejected & No & No & No & No & No & No & No & No \\ \bottomrule
\end{tabular}
\end{table}
%(end)

The null hypothesis evaluated with this test checks if at least two of the
samples represent populations with different median values, in a set of $k$
samples (where $k \ge 2$). 
If the null hypothesis is rejected, then the quality of the results generated
by the proposed MMAS variants is not equivalent.
Table~\ref{tab:stat-test} shows the results of the Friedman test for the six
proposed MMAS variants: MMAS-RWM-LC, MMAS-RWM-BT, MMAS-RWM-CT, MMAS-WRS-LC,
MMAS-WRS-BT and MMAS-WRS-CT.
The test was calculated separately for each of the eight TSP instances
investigated.
As can be seen from Table~\ref{tab:stat-test}, the null hypothesis was not
rejected in any of the cases, hence no statistically significant differences in
the quality of the results generated by the MMAS variants were found.

\subsection{Local Search}
\label{sec:Local_Search}

The ACO algorithms are typically combined with problem-specific LS
heuristics~\cite{Dorigo2004}. In this combination, the ACO performs
a \emph{coarse-grained} search throughout the space of possible solutions,
while the LS is responsible for a fine-grained exploitation of the neighborhood
of a solution built by an ant.

For the sake of completeness of the current work, we have implemented a parallel version of the
2-opt heuristic as described by Bentley~\cite{Bentley1992}.
The 2-opt works by searching for a pair of edges to remove so that
the resulting parts of the route can be reconnected using a new pair
of edges that has a combined length smaller than the length of the removed pair.
If such an improvement is found and applied, the
search is restarted until no further improvements can be found.
Although there are only $O(n^2)$ possible pairs to consider while searching for
the improvement, it is still too time-consuming if the number of nodes is on the
order of thousands or more. For this reason, in our implementation
the performance-oriented heuristics described by Bentley are applied, i.e. the search for a new
edge to insert is limited only to the nearest neighbors of the considered node (32 in our
case) and an array of so-called \emph{don't look bits} is used to further limit the
search. In the proposed parallel version of the 2-opt, a number of threads divided into
groups of 32 threads each (warps) search in parallel for the pair of edges to
remove, and, if any group succeeds, the improvement is applied (in
parallel) by all groups of threads reversing the respective part of the route.
The 2-opt is applied to every solution constructed by the ants.

It is worth mentioning, that the GPU-based implementation of the 2-opt
was first proposed by Rocki and Suda~\cite{Rocki2013} but without any additional
heuristics that could limit the $O(n^2)$ search space size. The approach was refined by
Zhou et al.~\cite{Zhou2016} who improved its execution speed and,
combined it with the Iterated Local Search. The authors solved TSP instances of up
to 4,461 nodes obtaining for the largest instance solutions with the mean
distance from the optimum equal to 3.67\%.
Another proposal to improve the 2-opt implementation by Rocki and Suda was made
by Robinson et al.~\cite{Robinson2018}, who allowed multiple non-conflicting
changes to a route to be performed simultaneously. The resulting 2-opt
hill-climbing with random restarts was used to solve the 7,397-city
instance with an error of around 8\% relative to the optimum.
The GPU-computing power was also utilized to solve the TSP in the work of Wang et
al.~\cite{Wang2017}, who proposed a parallel computation model for the self-organizing
map neural network. The authors considered instances up to 85,900 cities but the
quality of obtained solutions was typically at least 5\% above the respective optimum.

To assess the performance of the proposed parallel MMAS with the 2-opt, we
conducted a number of experiments using the Nvidia V100 GPU and several
instances from the TSPLIB repository with up to 18,512 nodes.
Specifically, the MMAS-RWM-BT with the candidate lists of length 32 was run
with the number of ants equal to 800 (multiple of the number of 80 SMs in the V100).
The values of the remaining parameters were set based on the size of the
instance. For the instances with up to $10^5$ nodes,
the number of iterations was 2,000 and the pheromone was evaporated slowly
($\rho = 0.9$), while for the larger
instances the number of iterations was increased to 3,000 and the speed of
the pheromone evaporation was higher \R{($\rho = 0.7$)}.
These values were selected with the aim of allowing the algorithm to find good
quality solutions within a relatively short time.
The computations were repeated 20 times.

%(fold)
\begin{table}[]
\footnotesize
\centering
\caption{\R{
    A comparison of the results generated by the ESACO~\cite{Ismkhan2017} and MMAS-RWM-BT
    (with the 2-opt LS) metaheuristics for the TSP instances from
    the TSPLIB repository. The best and mean solution costs (route lengths)
    are given along with the relative difference from the optimum reported in
    round brackets.  The values in the \emph{Time} column refer to the CPU runtime (in
    seconds) as reported in the respective work and the Nvidia Tesla V100 GPU
    runtime in the case of the MMAS-RWM-BT. The lowest mean cost for
    every instance is marked in bold.
}}
\R{
\label{tab:cmp-with-esaco}
\resizebox{\textwidth}{!}{%
\begin{tabular}{@{}lrrrrrrr@{}}
\toprule
\multicolumn{1}{c}{\multirow{2}{*}{Instance}} 
    & \multicolumn{1}{c}{\multirow{2}{*}{Optimum}} 
    & \multicolumn{3}{c}{ESACO\cite{Ismkhan2017}} 
    & \multicolumn{3}{c}{MMAS-RWM-BT} \\ \cmidrule(lr){3-5}  \cmidrule(l){6-8}
\multicolumn{1}{c}{} & \multicolumn{1}{c}{} & \multicolumn{1}{c}{Best solution} & \multicolumn{1}{c}{Mean solution} & \multicolumn{1}{c}{Time} & \multicolumn{1}{c}{Best solution} & \multicolumn{1}{c}{Mean solution} & \multicolumn{1}{c}{Time} \\ \midrule
\textit{eil51} & 426 & 426 (0\%) & \textbf{426.0 (0\%)} & 1.12 & 426 (0\%) & \textbf{426.0 (0\%)} & 0.43 \\
\textit{eil76} & 538 & 538 (0\%) & \textbf{538.0 (0\%)} & 1.39 & 538 (0\%) & \textbf{538.0 (0\%)} & 0.49 \\
\textit{kroA100} & 21282 & 21282 (0\%) & \textbf{21282.0 (0\%)} & 2.61 & 21282 (0\%) & \textbf{21282.0 (0\%)} & 0.58 \\
\textit{lin105} & 14379 & 14379 (0\%) & \textbf{14379.0 (0\%)} & 2.0 & 14379 (0\%) & \textbf{14379.0 (0\%)} & 0.6 \\
\textit{d198} & 15780 & 15780 (0\%) & \textbf{15780.0 (0\%)} & 6.5 & 15780 (0\%) & \textbf{15780.0 (0\%)} & 0.9 \\
\textit{kroA200} & 29368 & 29368 (0\%) & \textbf{29368.0 (0\%)} & 4.7 & 29368 (0\%) & \textbf{29368.0 (0\%)} & 0.8 \\
\textit{a280} & 2579 & 2579 (0\%) & 2579.1 (0\%) & 4.5 & 2579 (0\%) & \textbf{2579.0 (0\%)} & 1.0 \\
\textit{lin318} & 42029 & 42029 (0\%) & \textbf{42053.9 (0.06\%)} & 10.2 & 42029 (0\%) & 42069.6 (0.1\%) & 1.7 \\
\textit{pcb442} & 50778 & 50778 (0\%) & \textbf{50803.6 (0.05\%)} & 11.5 & 50809 (0.06\%) & 50950.7 (0.34\%) & 2.0 \\
\textit{att532} & 27686 & 27686 (0\%) & \textbf{27701.2 (0\%)} & 23.1 & 27686 (0\%) & 27708.9 (0.08\%) & 1.8 \\
\textit{rat783} & 8806 & 8806 (0\%) & \textbf{8809.8 (0.04\%)} & 22.6 & 8810 (0.05\%) & 8825.5 (0.22\%) & 3.1 \\
\textit{pr1002} & 259045 & 259045 (0\%) & \textbf{259509.0 (0.18\%)} & 35.8 & 259415 (0.14\%) & 259712.7 (0.26\%) & 4.0 \\
\textit{fl3795} & 28772 & 28787 (0.05\%) & 28883.5 (0.39\%) & 119.3 & 28793 (0.07\%) & \textbf{28819.3 (0.16\%)} & 19.5 \\
\textit{fnl4461} & 182566 & 183254 (0.38\%) & \textbf{183446.0 (0.48\%)} & 192.6 & 183361 (0.44\%) & 183627.6 (0.58\%) & 34.9 \\
\textit{rl5915} & 565530 & 567177 (0.29\%) & 568935.0 (0.60\%) & 216.9 & 566123 (0.11\%) & \textbf{567699.9 (0.38\%)} & 49.3 \\
\textit{pla7397} & 23260728 & 23345479 (0.36\%) & 23389341.0 (0.55\%) & 213.9 & 23365046 (0.45\%) & \textbf{23386240.5 (0.54\%)} & 58.0 \\
\textit{rl11849} & 923288 & 928876 (0.61\%) & 930338.0 (0.76\%) & 575.8 & 926840 (0.38\%) & \textbf{928618.83 (0.58\%)} & 247.4 \\
\textit{usa13509} & 19982859 & 20172735 (0.95\%) & 20195089.0 (1.06\%) & 914.2 & 20128078 (0.73\%) & \textbf{20168030.2 (0.93\%)} & 224.2 \\
\textit{brd14051} & 469385 & 473718 (0.92\%) & \textbf{474087.0 (1.00\%)} & 682.5 & 473389 (0.85\%) & 474715.65 (1.14\%) & 228.4 \\
\textit{d15112} & 1573084 & 1587150 (0.89\%) & 1589288.0 (1.03\%) & 776.7 & 1584054 (0.70\%) & \textbf{1586604.05 (0.86\%)} & 404.2 \\
\textit{d18512} & 645238 & 652516 (1.13\%) & 653154.0 (1.23\%) & 684.4 & 650784 (0.86\%) & \textbf{651413.58 (0.96\%)} & 335.5 \\ \bottomrule
\end{tabular}%
}
} %~\R
\end{table}
%(end)

%(fold)
\begin{table}[]
\footnotesize
\centering
\R{
\caption{\R{
    A comparison of the results generated by the MFC-ABC and MMAS-RWM-BT
    (with the 2-opt LS) metaheuristics. The meaning of the columns is the same
    as in Tab.~\ref{tab:cmp-with-esaco}.
}}
\label{tab:cmp-with-mfc-abc}

\begin{tabular}{@{}lrrrrrr@{}}
\toprule
\multicolumn{1}{c}{\multirow{2}{*}{Instance}} 
    & \multicolumn{1}{c}{\multirow{2}{*}{Optimum}} 
    & \multicolumn{2}{c}{MCF-ABC\cite{Choong2019}} 
    & \multicolumn{3}{c}{MMAS-RWM-BT} \\ \cmidrule(lr){3-4}  \cmidrule(l){5-7}
    \multicolumn{1}{c}{} & \multicolumn{1}{c}{} & \multicolumn{1}{c}{Mean solution} & \multicolumn{1}{c}{Time} & \multicolumn{1}{c}{Best solution} & \multicolumn{1}{c}{Mean solution} & \multicolumn{1}{c}{Time} \\ \midrule
\textit{eil101} & 629 & \textbf{629.0 (0\%)} & 0.0 & 629 (0\%) & \textbf{629.0 (0\%)} & 0.6 \\
\textit{lin105} & 14379 & \textbf{14379.0 (0\%)} & 0.0 & 14379 (0\%) & \textbf{14379.0 (0\%)} & 0.6 \\
\textit{pr107} & 44303 & \textbf{44303.0 (0\%)} & 0.0 & 44303 (0\%) & \textbf{44303.0 (0\%)} & 0.6 \\
\textit{gr120} & 6942 & \textbf{6942.0 (0\%)} & 0.0 & 6942 (0\%) & \textbf{6942.0 (0\%)} & 0.6 \\
\textit{pr124} & 59030 & \textbf{59030.0 (0\%)} & 0.1 & 59030 (0\%) & \textbf{59030.0 (0\%)} & 0.6 \\
\textit{bier127} & 118282 & \textbf{118282.0 (0\%)} & 0.1 & 118282 (0\%) & \textbf{118282.0 (0\%)} & 0.7 \\
\textit{ch130} & 6110 & \textbf{6110.0 (0\%)} & 0.0 & 6110 (0\%) & \textbf{6110.0 (0\%)} & 0.6 \\
\textit{pr136} & 96772 & \textbf{96772.0 (0\%)} & 0.2 & 96772 (0\%) & \textbf{96772.0 (0\%)} & 0.6 \\
\textit{gr137} & 69853 & \textbf{69853.0 (0\%)} & 0.1 & 69853 (0\%) & \textbf{69853.0 (0\%)} & 0.7 \\
\textit{pr144} & 58537 & \textbf{58537.0 (0\%)} & 1.8 & 58537 (0\%) & \textbf{58537.0 (0\%)} & 0.6 \\
\textit{ch150} & 6528 & \textbf{6528.0 (0\%)} & 0.1 & 6528 (0\%) & \textbf{6528.0 (0\%)} & 0.7 \\
\textit{kroA150} & 26524 & \textbf{26524.0 (0\%)} & 0.1 & 26524 (0\%) & \textbf{26524.0 (0\%)} & 0.7 \\
\textit{kroB150} & 26130 & \textbf{26130.0 (0\%)} & 0.1 & 26130 (0\%) & \textbf{26130.0 (0\%)} & 0.7 \\
\textit{pr152} & 73682 & \textbf{73682.0 (0\%)} & 1.8 & 73682 (0\%) & \textbf{73682.0 (0\%)} & 0.8 \\
\textit{u159} & 42080 & \textbf{42080.0 (0\%)} & 0.1 & 42080 (0\%) & \textbf{42080.0 (0\%)} & 0.7 \\
\textit{si175} & 21407 & \textbf{21407.0 (0\%)} & 0.2 & 21407 (0\%) & 21407.5 (0\%) & 0.8 \\
\textit{brg180} & 1950 & \textbf{1950.0 (0\%)} & 0.0 & 1950 (0\%) & \textbf{1950.0 (0\%)} & 0.7 \\
\textit{rat195} & 2323 & \textbf{2323.0 (0\%)} & 0.2 & 2323 (0\%) & 2325.2 (0.2\%) & 0.8 \\
\textit{d198} & 15780 & \textbf{15780.0 (0\%)} & 1.3 & 15780 (0\%) & \textbf{15780.0 (0\%)} & 0.9 \\
\textit{kroA200} & 29368 & \textbf{29368.0 (0\%)} & 0.1 & 29368 (0\%) & \textbf{29368.0 (0\%)} & 0.8 \\
\textit{kroB200} & 29437 & \textbf{29437.0 (0\%)} & 0.0 & 29437 (0\%) & \textbf{29437.0 (0\%)} & 0.8 \\
\textit{gr202} & 40160 & \textbf{40160.0 (0\%)} & 0.6 & 40160(0\%) & 40161.3(0\%) & 0.9 \\
\textit{tsp225} & 3916 & \textbf{3916.0 (0\%)} & 0.0 & 3916 (0\%) & \textbf{3916.0 (0\%)} & 0.8 \\
\textit{ts225} & 126643 & \textbf{126643.0 (0\%)} & 0.1 & 126643 (0\%) & \textbf{126643.0 (0\%)} & 0.9 \\
\textit{pr226} & 80369 & \textbf{80369.0 (0\%)} & 1.4 & 80369 (0\%) & \textbf{80369.0 (0\%)} & 0.9 \\
\textit{gr229} & 134602 & \textbf{134602.0 (0\%)} & 0.8 & 134602 (0\%) & 134608.4 (0\%) & 0.9 \\
\textit{gil262} & 2378 & \textbf{2378.0 (0\%)} & 0.1 & 2378 (0\%) & 2378.2 (0.01\%) & 1.0 \\
\textit{pr264} & 49135 & \textbf{49135.0 (0\%)} & 0.1 & 49135 (0\%) & \textbf{49135.0 (0\%)} & 1.0 \\
\textit{a280} & 2579 & \textbf{2579.0 (0\%)} & 0.0 & 2579 (0\%) & \textbf{2579.0 (0\%)} & 1.0 \\
\textit{pr299} & 48191 & \textbf{48191.0 (0\%)} & 0.2 & 48191 (0\%) & 48197.1 (0.01\%) & 1.1 \\
\textit{lin318} & 42029 & \textbf{42029.0 (0\%)} & 1.5 & 42029 (0\%) & 42069.6 (0.1\%) & 1.7 \\
\textit{rd400} & 15281 & \textbf{15281.0 (0\%)} & 2.4 & 15281 (0\%) & 15282.9 (0.01\%) & 1.4 \\
\textit{fl417} & 11861 & \textbf{11861.0 (0\%)} & 5.7 & 11861 (0\%) & \textbf{11861.0 (0\%)} & 1.6 \\
\textit{gr431} & 171414 & \textbf{171414.0 (0\%)} & 13.0 & 171414 (0\%) & 171419.7 (0\%) & 1.5 \\
\textit{pr439} & 107217 & \textbf{107217.0 (0\%)} & 2.3 & 107217 (0\%) & \textbf{107217.0 (0\%)} & 1.4 \\
\textit{pcb442} & 50778 & \textbf{50778.0 (0\%)} & 1.6 & 50809 (0.06\%) & 50950.7 (0.34\%) & 2.0 \\
\textit{d493} & 35002 & \textbf{35002.7 (0\%)} & 20.9 & 35004 (0.01\%) & 35046.9 (0.13\%) & 1.7 \\
\textit{att532} & 27686 & \textbf{27686.5 (0\%)} & 11.4 & 27686 (0\%) & 27708.9 (0.08\%) & 1.8 \\
\textit{ali535} & 202339 & \textbf{202339.0 (0\%)} & 10.6 & 202339 (0\%) & 202378.1 (0.02\%) & 2.0 \\
\textit{si535} & 48450 & 48498.3 (0.1\%) & 53.3 & 48450 (0\%) & \textbf{48455.7 (0.01\%)} & 1.9 \\
\textit{pa561} & 2763 & \textbf{2763.1 (0\%)} & 7.1 & 2763 (0\%) & 2768.4 (0.20\%) & 1.8 \\
\textit{u574} & 36905 & \textbf{36905.0 (0\%)} & 2.1 & 36905 (0\%) & 36923.2 (0.05\%) & 2.1 \\
\textit{rat575} & 6773 & \textbf{6774.3 (0.02\%)} & 6.3 & 6774 (0.01\%) & 6781.5 (0.13\%) & 1.9 \\
\textit{p654} & 34643 & \textbf{34643.0 (0\%)} & 21.5 & 34643 (0\%) & 34653.6 (0.03\%) & 2.6 \\
\textit{d657} & 48912 & \textbf{48915.1 (0.01\%)} & 15.5 & 48913 (0\%) & 48993.2 (0.17\%) & 2.3 \\
\textit{gr666} & 294358 & \textbf{294404.8 (0.02\%)} & 32.1 & 294358 (0\%) & 294544.3 (0.06\%) & 2.2 \\
\textit{u724} & 41910 & \textbf{41916.5 (0.02\%)} & 12.9 & 41929 (0.05\%) & 41973.8 (0.15\%) & 2.3 \\
\textit{rat783} & 8806 & \textbf{8806.0 (0\%)} & 4.2 & 8810 (0.05\%) & 8825.5 (0.22\%) & 3.1 \\
\textit{dsj1000} & 18659688 & \textbf{18661580.6 (0.01\%)} & 53.2 & 18665058 (0.03\%) & 18677500.3 (0.10\%) & 4.4 \\
\textit{pr1002} & 259045 & \textbf{259073.0 (0.01\%)} & 16.6 & 259415 (0.14\%) & 259712.7 (0.26\%) & 4.0 \\
\textit{si1032} & 92650 & \textbf{92650.0 (0\%)} & 4.4 & 92650 (0\%) & \textbf{92650.0 (0\%)} & 3.3 \\
\textit{vm1084} & 239297 & \textbf{239322.3 (0.01\%)} & 24.3 & 239297 (0\%) & 239480.0 (0.08\%) & 3.9 \\
\textit{pcb1173} & 56892 & \textbf{56897.9 (0.01\%)} & 15.4 & 56892 (0\%) & 56994.7 (0.18\%) & 3.8 \\
\textit{d1291} & 50801 & \textbf{50843.7 (0.08\%)} & 15.7 & 50803 (0\%) & 50855.5 (0.11\%) & 4.5 \\
\textit{d1655} & 62128 & \textbf{62221.6 (0.15\%)} & 27.0 & 62192 (0.10\%) & 62321.1 (0.31\%) & 5.7 \\
\textit{u1817} & 57201 & \textbf{57354.2 (0.27\%)} & 24.6 & 57209 (0.01\%) & 57428.2 (0.4\%) & 7.0 \\
\textit{u2152} & 64253 & \textbf{64426.5 (0.27\%)} & 28.3 & 64366 (0.18\%) & 64520.1 (0.42\%) & 7.1 \\
\textit{pr2392} & 378032 & \textbf{378549.6 (0.14\%)} & 28.9 & 378390 (0.1\%) & 379872.0 (0.49\%) & 10.3 \\
\textit{fl3795} & 28772 & 28825.1 (0.18\%) & 131.7 & 28793 (0.07\%) & \textbf{28819.3 (0.16\%)} & 19.5 \\
\textit{fnl4461} & 182566 & \textbf{183002.8 (0.24\%)} & 66.6 & 183361 (0.44\%) & 183627.6 (0.58\%) & 34.9 \\
\textit{rl5915} & 565530 & 567990.2 (0.44\%) & 94.2 & 566123 (0.11\%) & \textbf{567699.9 (0.38\%)} & 49.3 \\
\textit{pla7397} & 23260728 & \textbf{23324321.9 (0.24\%)} & 247.6 & 23365046 (0.45\%) & 23386240.5 (0.54\%) & 58.0 \\
\textit{rl11849} & 923288 & \textbf{928015.1 (0.51\%)} & 308.9 & 926840 (0.38\%) & 928618.83 (0.58\%) & 247.4 \\ \bottomrule
\end{tabular}
}%~\R
\end{table}
%(end)

\R{
Table~\ref{tab:cmp-with-esaco} and Tab.~\ref{tab:cmp-with-mfc-abc} present the
obtained results and compare them with the
results of two high-performing metaheuristics, namely the Effective Strategies+ACO
(ESACO)~\cite{Ismkhan2017} and the Artificial Bee Colony with a Modified Choice
Function (MFC-ABC)~\cite{Choong2019} (respectively). The ESACO was chosen as it is the
current state-of-the-art ACO-based TSP solver. Specifically, it is based on the
ACS combined with an efficient LS comprising the 2-opt, 3-opt and so-called
\emph{double-bridge moves}. The second algorithm, the MFC-ABC, combines a modified ABC
metaheuristic with the well-known Lin-Kernighan heuristic.  Extensive comparisons with
the existing state-of-the-art metaheuristic TSP solvers showed that the MFC-ABC is
very competitive in terms of both computation time and quality of the
results~\cite{Choong2019}.

Analysis of the results presented in Tab.~\ref{tab:cmp-with-esaco} and
Tab.~\ref{tab:cmp-with-mfc-abc} reveals
several facts.
Firstly, the proposed MMAS with the 2-opt LS is able to obtain
good quality results, i.e. within 1\% from the optima, for all but one
instance, \emph{brd14051}, for which the mean cost was 1.14\% above the optimum.
As expected, the error increases with the instance size.
Secondly, the 2-opt LS is too simplistic to allow the optima to be found except
for small instances. On the other hand, it proved to be effective
enough to allow the proposed implementation to obtain better results than the
ESACO in 14 out of 21 cases, including the largest instances with up to
18,512 cities. 

A comparison of the proposed GPU-based MMAS with the MFC-ABC (Tab.~\ref{tab:cmp-with-mfc-abc}) confirms the effectiveness of the latter as it
generated (on average) better results for 33 out of 63 (52\%) instances while
worse only for 3. Still, the results obtained for the
GPU-based MMAS can be considered satisfactory with the relative error being
below 0.5\% for all but three (\emph{fnl4461, pla7397} and
\emph{rl11849}) instances for which it reached 0.58\%.
Unfortunately, the $O(n^2)$ memory complexity of the MMAS prevents solving
the largest TSP instance (with 85,900 cities) considered in~\cite{Choong2019} 
as it would require more than 16 GB of RAM offered by the V100 GPU.

It is also worth adding that the GPU-based MMAS is also
competitive in terms of computation time.
The average runtime for the instances with less than 10,000 cities was
below 1 minute, whereas solving each of the remaining (larger) instances took less than
7 minutes.

Summarizing, the results show that the high computational power of the GPU
is enough to allow the relatively simple method (MMAS with the 2-opt) to compete
with sophisticated but sequential algorithms. This suggests that more effort
should be put into the development of parallel versions of the more advanced
LS methods, e.g., the Lin-Kernighan heuristic, allowing
the available computing power of multi-core CPUs and the increasingly
popular GPUs to be fully utilized.
}

\section{Conclusions}
\label{sec:Conclusions}

\label{sec:conclusions}

The efficient parallel execution of the ACO on GPUs requires a careful
organization of the computations, which must harmonize with their capabilities
and limitations.
A large number of processing elements (cores) allow a high
degree of parallelism, and thus require/benefit-from high concurrency.
In this paper, we have presented a novel parallel
implementation of the solution construction procedure, which is, in general, the most
important and the most time-consuming part of the MMAS and ACO.
Specifically, we have proposed a novel implementation of the node selection
procedure based on the WRS algorithm~\cite{Efraimidis2006}.  It requires little
cooperation between parallel threads, resulting in a high degree of
scalability, yet it is easy to implement and offers the same quality of
generated solutions as does the original RWM-based method.
The efficient use of the fast, but size limited, shared memory offered by the
SMs that comprise the GPU is also essential from the performance point of view.
We have discussed three tabu implementations, namely, LC, CT, and BT, which
differ in memory overhead and the time complexity of the offered operations.
A total of six MMAS
variants have been evaluated empirically on a set of TSP instances ranging
from 198 to 3,795 cities.
The computations were conducted for two subsequent Nvidia
GPU architectures, namely, Pascal (P100) and Volta (V100).
In general, the MMAS with the parallel WRS-based node selection implementation
is faster than the MMAS with a parallel version of the RWM-based method.
The impact of the tabu implementation depends on the size of the problem
instance and, more importantly, on the use of the candidate lists.
If the candidate lists are
used, the BT is preferred as the slow enumerations of its contents are
executed rarely.
On the other hand, if the lists are not used, then the proposed CT
offers better speed at the cost of increased shared memory usage.
Overall, the MMAS-WRS-BT and MMAS-WRS-CT are the recommended choices,
depending on whether the candidate lists are used or not, respectively.
Both GPUs offer excellent performance, with the newer V100 GPU being up to 2.88x
faster than the previous-generation P100.
For example, the solution construction phase for the \emph{pr1002} instance
takes about 6.09 ms on the P100 GPU and only 2.16 on the V100 GPU.
If candidate lists with a size of 32 are used, the MMAS-WRS-BT executing on the
V100 needs only 0.93 ms, i.e., it generates solutions at the impressive rate of
over 1 million solutions per second.

The analysis of the results presented in the literature on speeding 
up the ACO's execution using the GPU-based computing shows that
huge progress has been made since the emergence of GPUs as a platform
for general-purpose computing.
For example, it took approximately 392 ms per iteration of the ACO to
build solutions to the \emph{pr1002} TSP instance when running 
on the C2050 (Fermi) GPU~\cite{Cecilia2012}.
Five GPU generations forward, and the time has decreased to about 2 ms
for the V100 (Volta) GPU, that is, 196x faster.
Overall, advances in the design of GPU architecture,
progress in the development of software tools (compilers, libraries),
and algorithmic refinements seem to be a promising answer to the slowing-down 
of Moore's law.
Additionally, the computational experiments considering the MMAS combined with
the 2-opt LS, show that the high computing speed of GPUs may be sufficient to
compete with more sophisticated but sequential algorithms, especially
if the results are to be generated quickly.

\subsection*{Future work}

The presented work may be extended in multiple directions.  In this paper, we
have considered TSP instances for which the tabu fits entirely into the shared
(local) memory of a group of threads.  However, its size is very limited, up to
64kB or 96kB even for newer Nvidia GPU architectures (Pascal, Volta).  If part
of the shared memory is reserved for other purposes, for example, a local
search procedure, this leaves even less space for the tabu.  It may be
interesting to consider tabu implementations in which the shared memory is used
along with global memory, i.e., variants of the described BT.  It is worth
adding, that the proposed implementations of the tabu data structure and the
WRS-based node selection procedure can be applied to other ACO-based
algorithms, for example, the ACS~\cite{Skinderowicz2016}, and even to other
metaheuristics.

In this present work, we have mostly focused on the algorithmic refinements of
the GPU-based ACO implementation, but we think that further improvements are
possible. We agree with Cecilia et al.~\cite{Cecilia2018} that the new
mechanism of cooperative groups introduced in CUDA 9 could prove useful.  Also,
improvements to the SIMT model introduced in the Volta architecture are worth
investigating as they allow for more divergence in the computations performed
by the threads belonging to the same warp~\cite{V1002018}.
Additionally, solving large problem instances would require significant changes
to the MMAS to be made, including a replacement of the pheromone memory with a more
space-efficient alternative and exploring ideas for alternative, faster
proportional selection method implementations.

\noindent
\textbf{Acknowledgments:} This research was supported in part by PL-Grid Infrastructure.

\bibliographystyle{plainnat}
\footnotesize
\bibliography{article}

% Add a bibliography block to the postdoc
\end{document}